\documentclass[sigconf,authorversion=true]{acmart}

\usepackage{booktabs} % For formal tables
\usepackage{graphicx}
\usepackage{subfig}
\usepackage[linesnumbered, vlined, ruled]{algorithm2e}
\usepackage{tablefootnote}
\usepackage[font=small,skip=0.5em]{caption}

\def\vector#1{\mbox{\boldmath $#1$}}
\newcommand{\CR}{C}

\copyrightyear{2017}
\acmYear{2017}
\setcopyright{acmcopyright}
\acmConference{GECCO '17}{July 15-19, 2017}{Berlin, Germany}\acmPrice{15.00}\acmDOI{http://dx.doi.org/10.1145/3071178.3071226}
\acmISBN{978-1-4503-4920-8/17/07}

\begin{document}
\title[TPAM: A Simulation-Based Model]{TPAM: A Simulation-Based Model for Quantitatively Analyzing Parameter Adaptation Methods}

%% \title{SIG Proceedings Paper in LaTeX Format}
%% \titlenote{Produces the permission block, and
%%   copyright information}
%% \subtitle{Extended Abstract}
%% \subtitlenote{The full version of the author's guide is available as
%%   \texttt{acmart.pdf} document}

\author{Ryoji Tanabe}
%\authornote{Dr.~Trovato insisted his name be first.}
\affiliation{%
%  \institution{Department of Computer Science and Engineering}
  \institution{Southern University of Science and Technology}
%  \streetaddress{\hspace{1em}}
%  \city{Shenzhen, China} 
%  \state{\hspace{1em}} 
%  \postcode{\hspace{1em}}
}
\email{rt.ryoji.tanabe@gmail.com}

\author{Alex Fukunaga}
%\authornote{Dr.~Trovato insisted his name be first.}
\affiliation{%
%  \institution{Institute of Space and Astronautical Science}
  \institution{The University of Tokyo}
%  \streetaddress{\hspace{1em}}
 % \city{Tokyo, Japan} 
%  \state{\hspace{1em}} 
%  \postcode{\hspace{1em}}
}
\email{fukunaga@idea.c.u-tokyo.ac.jp}

\begin{abstract}
While a large number of adaptive Differential Evolution (DE) algorithms have been proposed, their Parameter Adaptation Methods (PAMs) are not well understood.
We propose a Target function-based PAM simulation (TPAM) framework for evaluating the tracking performance of PAMs.
The proposed TPAM simulation framework measures the ability of PAMs to track predefined target parameters, 
thus enabling quantitative analysis of the adaptive behavior of PAMs.
We evaluate the tracking performance of PAMs of widely used five adaptive DEs (jDE, EPSDE, JADE, MDE, and SHADE) on the proposed TPAM, and
show that TPAM can provide important insights on PAMs, e.g., why the PAM of SHADE performs better than that of JADE, and under what conditions the PAM of EPSDE fails at parameter adaptation.

%We propose a Target function-based simulation framework for evaluating the tracking performance of PAMs (TPAM).

%In particular, as far as we know, there is no previous work investigating the parameter adaptation ability of PAMs in adaptive DE.

\end{abstract}



% We no longer use \terms command


\maketitle

\section{Introduction}
\label{sec:introduction}

%% PriceSL05
%% StornP97

Continuous black-box optimization is the problem of finding a $D$-dimensional solution $\vector{x} = (x_1, ..., x_D)^{\rm T} \in \mathbb{R}^D$  that minimizes an objective function $f: \mathbb{R}^D \rightarrow \mathbb{R}$ without explicit knowledge of the form or structure of $f$.
Differential Evolution (DE) is one of most efficient Evolutionary Algorithms (EAs) for continuous optimization \cite{StornP97}, and has been applied to many real-world problems \cite{DasMS16}.
%
%

%GamperleMK02, RonkkonenKP05, ZielinskiWLK06, 

While the fact that the search performance of EAs is strongly influenced by its control parameter settings has been widely accepted in the evolutionary computation community for decades \cite{EibenHM99,KarafotiasHE15}, it was initially reported that the search performance of DE was fairly robust with respect to control parameter settings \cite{StornP97}.
However, later work showed that in fact, the performance of DE was significantly affected by control parameter settings \cite{BrestGBMZ06}.
As a result, research in automated parameter control methods for DE has become an active area of research since around 2005. 
In recent years, the DE community has focused on adaptive control parameter methods \cite{DasMS16} which adjust control parameters online during search.
Some representative adaptive DEs are jDE \cite{BrestGBMZ06}, JADE \cite{ZhangS09}, EPSDE \cite{MallipeddiSPT11}, MDE \cite{IslamDGRS12}, and SHADE \cite{TanabeF13}.
Almost all adaptive DEs adjust two control parameters: the scale factor $F \in [0, 1)$ and the crossover rate $\CR \in [0,1]$ (For details, see Section \ref{sec:ade}).

%\cite{BrestGBMZ06, ZhangS09, MallipeddiSPT11, BrestZBGZ08, Dick10, SeguraCSL14, DrozdikAAT15, SeguraCSL15,ZamudaB15}.

However, while many adaptive DEs have been proposed, their {\bf P}arameter {\bf A}daptation {\bf M}ethods ({\bf PAM}s) are poorly understood.
Previous work on adaptive DEs such as  \cite{BrestGBMZ06,MallipeddiSPT11,ZhangS09,IslamDGRS12,TanabeF13} has tended to propose a novel adaptive DE variant and evaluate its performance on some benchmark functions, but analysis of their  adaptation methods have been minimal.
% as follows: {\em Similarly to the main body of work within EC, most publications about a parameter control mechanism simply show that it works. Usually, there is no or very little discussion about why it works and how it works.}
Several previous works have tried to analyze PAMs in adaptive DE \cite{BrestZBGZ08,ZielinskiWL08,ZhangS09,DrozdikAAT15,SeguraCSL14,ZamudaB15}.
However, these previous analyses have been mostly limited to plots of  changes in $F$ and $\CR$ values during a typical run on  benchmark functions, and the analysis has been limited to qualitative descriptions such as  ``in this adaptive DE the meta-parameter of $\CR$ quickly drops down to $[\CR_1, \CR_2]$ after several iterations on benchmark function $f_1$, while it gradually increases to $[\CR_3, \CR_4]$ on benchmark function $f_2$''.
This previous approach (plotting parameter values) is fundamentally limited because they can only lead to very weak, qualitative conclusions of the form:
(1) ``for a given problem, parameter values for a given PAM depend on the current state of the search''
(2) ``different PAMs lead to different parameter trajectories''
(3) ``the parameter trajectory of a given PAM is problem-dependent''.
In other words, the behavior and limitations of PAMs for DE are currently poorly understood, and
previous analyses have not yielded significant insights into fundamental questions such as: ``{\bf why} does PAM1 perform better than PAM2 on a given problem?''.
This situation is not unique to the DE community -- Karafotias et al. \cite{KarafotiasHE15} have pointed out the lack of the analysis of adaptation mechanisms in EAs. 
For example, even in the field of Evolution Strategies (ESs) \cite{HansenAA15}, where step size adaptation has been studied since the earliest days of the field of evolutionary computation, such adaptation mechanisms are far from being well-understood \cite{HansenAA14}.

In fact, in previous work, the crucial term {\it adaptation} tends not to be clearly defined at all, which leaves one with little alternative but to compare search algorithm performance (as a proxy for how well the proposed adaptive mechanism works \cite{ZielinskiWL08,DrozdikAAT15,SeguraCSL14}).
It is difficult to define metrics for adaptation that can be applied to a wide range of control parameter adaptation mechanisms. 
Although some studies propose alternative metrics (e.g., the number of improvements \cite{SeguraCSL14,ZamudaB15}) to analyze PAMs, they cannot directly investigate PAMs and do not provide sufficient information.

One possible approach to quantitatively analyzing adaptation is to compare the control parameter values generated by a PAM to an ``optimal'' parameter value schedule.
However, in general, such theoretical, optimal parameter schedules are difficult to obtain and only known for very simple functions \cite{Back93,HansenAA15}.
%However, in general, such theoretical, optimal parameter schedules are difficult to obtain, even for very simple fitness functions.
%{\bf [XXX citeable paper on optimal parameter adaptation?]}
A recent simulation-based approach seeks to approximate optimal adaptive behavior \cite{TanabeF16}, but this is computationally very expensive and so far has been limited to a 1-step, greedy approximation. Furthermore, although comparisons of a PAM vs. optimal adaptive processes can allow an evaluation of the resulting search performance, such an approach does not necessarily yield insights that allows to understand why one PAM generates parameter adaptation histories closer to an optimal parameter adaptation schedule than another.
Thus, it seems that there are significant obstacles to analyzing  parameter adaptation as the problem of generating parameter trajectory which matches  a static, {\em a posteriori}  optimal parameter history.

{\bf In this paper, we take another approach which treats parameter adaptation as a problem of adapting to a dynamic environment which is constantly changing. More specifically, we propose a novel, empirical  model which treats the control parameter values modified by the PAM  (in the case of DE, the $F$ and $\CR$ values) as the ``output'' of the PAM, where this output is evaluated by comparison against a prespecified ``target'' function which changes over time, i.e., we assess PAMs by measuring how well they generate control parameter values which track a given, ``target function''.}  % statement of hypothesis/approach

We propose TPAM (Target function-based PAM simulation), a simulation based approach to analyzing the behavior of PAMs
which measures how effectively a PAM is able to track a given, ideal ``target'' function\footnote{Of course, the proposed TPAM can simulate a nondynamic environment using a target function such as $g (n_t) = 0.5$ and approximate optimal parameter adaptation process which is experimentally obtained by GAO \cite{TanabeF16}. See Section \ref{sec:tpam_tf}.}.
Note that this paper focuses on  {\em parameter adaptation methods of adaptive DEs} for $F$ and $\CR$, {\em not adaptive DEs} as in \cite{TanabeF16}.
In general, the term ``adaptive DE'' denotes a complex algorithm composed of multiple algorithm components.
For example, ``L-SHADE'' \cite{TanabeF14CEC} consists of four key components: (a) current-to-$p$best/1 mutation strategy, (b) binomial crossover, (c) the ``SHADE method'' for adapting  parameters  $F$ and $\CR$ (i.e., PAM-SHADE), and (d) linear population size reduction strategy.
In this paper, we are not interested in ``L-SHADE'', the complex DE algorithm composed of (a), (b),  (c), and (d) -- we want to focus on analyzing (c), {\it the PAM,  in isolation}.
Therefore, we extracted and studied only the PAM from each adaptive DE variant for our study. %, and generalized it so that it can be combined with arbitrary mutation and crossover methods.
%This approach follows  recent work on the DE community \cite{ZielinskiWL08,DrozdikAAT15,SeguraCSL15} as well as other adaptive EA variants \cite{FialhoSS09,PellegriniSB12}. 
%
Although many PAMs have been proposed in the literature, to our knowledge, there has been no previous work which analyzed the behavior of PAMs in isolation.
Our TPAM approach defines an ideal target trajectory and then performs a simulation which measures 
how closely a PAM tracks this target trajectory. 
This allows us to ask: ``how much better is PAM1 vs. PAM2 with respect to tracking a target control parameter trajectory?'', i.e., our approach 
 enables a {\it quantitative} comparison of the behavior of different PAMs, which yields new insights into {\em why} some PAMs lead to better DE performance than others.

%% This paper is organized as follows: Section \ref{sec:ade} introduces the PAMs of the five adaptive DEs.
%% Then, we describe the proposed TPAM framework in Section \ref{sec:proposed_method}.
%% We evaluate and analyze the five PAMs by using the TPAM framework in Section \ref{sec:experiment}.
%% Finally, Section \ref{sec:conclusion} concludes this paper and discusses our future work.

%\section{Parameter Adaptation Methods in Adaptive Differential Evolution}
\section{PAMs in adaptive DE}
%\section{Adaptive Differential Evolution}
\label{sec:ade}

This section first provides a brief overview of DE \cite{StornP97} and then reviews five PAMs in adaptive DE.

%\subsection{DE}

In DE, a population $\vector{P} = \{ \vector{x}^{1}, ..., \vector{x}^{N} \}$ is represented as a set of real parameter vector $\vector{x}^{i} = (x^{i}_{1}, ..., x^{i}_{D})^\mathrm{T}$, $i \in \{1, ..., N\}$, where $N$ is the population size.
After initialization of the population, for each iteration $t$, for each $\vector{x}^{i,t}$, a mutant vector $\vector{v}^{i,t}$ is generated from the individuals in $\vector{P}^t$ by applying a mutation strategy.
The most commonly used  mutation strategy is rand/1: $\vector{v}^{i,t} = \vector{x}^{r_1,t} + F_{i,t} \: (\vector{x}^{r_2,t} - \vector{x}^{r_3,t})$.
The indices $r_1$, $r_2$, $r_3$ above are randomly selected from $\{1, ..., N\} \backslash \{i\}$ such that they differ from each other.
The scale factor $F_{i,t} \in (0,1]$ controls the magnitude of the differential mutation operator.

Then, the mutant vector $\vector{v}^{i,t}$ is crossed with the parent $\vector{x}^{i,t}$ in order to generate a trial vector $\vector{u}^{i,t}$.
Binomial crossover, the most commonly used crossover method in DE, is implemented as follows: For each $j \in \{1, ..., D\}$, if ${\rm rand}[0,1] \leq \CR_{i,t}$ or $j = j_{r}$
(where, ${\rm rand[0,1]}$ denotes a uniformly generated random number from $[0, 1]$, and $j_r$ is a decision variable index which is uniformly randomly selected from $\{1, ..., D\}$),
then $u^{i,t}_j = v^{i,t}_{j}$. Otherwise, $u^{i,t}_j = x^{i,t}_{j}$.
$\CR_{i,t} \in [0,1]$ is the crossover rate.

After all of the trial vectors $\vector{u}^{i,t}$, $i \in \{1, ..., N\}$ have been generated, each individual $\vector{x}^{i,t}$ is compared with its corresponding trial vector $\vector{u}^{i,t}$, keeping the better individual in the population $\vector{P}^t$, i.e., if $f(\vector{u}^{i,t}) \leq f(\vector{x}^{i,t})$, $\vector{x}^{i,t+1} = \vector{u}^{i,t}$. Otherwise, $\vector{x}^{i,t+1} = \vector{x}^{i,t}$.
%

%\subsection{PAMs}

Five representative, adaptive DE algorithms which adapt 
the scale factor $F$ and the crossover rate $\CR$ are
jDE \cite{BrestGBMZ06}, EPSDE \cite{MallipeddiSPT11}, JADE \cite{ZhangS09}, MDE \cite{IslamDGRS12}, and SHADE \cite{TanabeF13}.
See Algorithm S.1 -- S.5 in the supplementary material for complete descriptions of the five PAMs described below:

%% See Algorithm $S.1 \sim S.5$ in the supplementary material for complete descriptions of these algorithms which describe components other than their adaptation mechanisms.

\begin{definition}{Trial vector success/failure}  We say that a generation of a trial vector is {\em successful} if $ f(\vector{u}^{i,t}) \leq f(\vector{x}^{i,t})$.
Otherwise, we say that the trial vector generation is a {\em failure}.
\label{def:success-failure}
\end{definition}

%% \subsection{jDE}
%% \label{sec:jde}

%{\bf [XXX:TODO  later sections use the PAM$^{name}$ instead of PAM-name. Make consistent]}

\noindent {\bf $\bullet$ PAM-jDE}:
A PAM in jDE \cite{BrestGBMZ06} assigns a different set of parameter values $F_{i,t}$ and $\CR_{i,t}$ to each $\vector{x}^{i,t}$ in $\vector{P}^t$.
For $t=1$, the parameters for all individuals $\vector{x}^{i,t}$ are set to $F_{i,t} = 0.5$ and $\CR_{i,t} = 0.9$.
In each iteration $t$, each parameter is randomly modified (within a pre-specified range) with some probability:
\begin{align}
\label{eqn:jde-f}
\small
%\footnotesize
F'_{i,t} &= \begin{cases}
{\rm rand}[0.1,1] &   {\rm if} \: {\rm rand}[0,1] < \tau_F\\
F_{i,t} &   {\rm otherwise}
  \end{cases}
\\
\label{eqn:jde-cr}
\small
%\footnotesize
\CR'_{i,t} &= \begin{cases}
{\rm rand}[0,1]  &   {\rm if} \: {\rm rand}[0,1] < \tau_{\CR}\\
\CR_{i,t} &   {\rm otherwise}
  \end{cases}
\end{align}
where $\tau_F$ and $\tau_{\CR} \in (0,1]$ are control parameters for parameter adaptation. % sentences shouldn't start with "Where" (with capital "W") -- it's acceptable (but better avoided)  to start with lower-case "where" when continuing a sentence before an equation definition.
Each individual $\vector{x}^{i,t}$ generates the trial vector using $F'_{i,t}$ and $\CR'_{i,t}$.
$F'_{i,t}$ and $\CR'_{i,t}$ are kept for the next iteration (i.e., $F_{i,t+1} = F'_{i,t}$ and $\CR_{{i,t+1}} = \CR'_{i,t}$) only when a trial is successful.

%% The overall generalized parameter adaptation method in jDE is described in Algorithm \ref{alg:general-jde} in the supplemental material.

%\subsection{EPSDE}

\noindent {\bf $\bullet$ PAM-EPSDE}:
PAM-EPSDE \cite{MallipeddiSPT11} uses an ``$F$-pool'' and a ``$\CR$-pool'' for parameter adaptation of $F$ and $\CR$,  respectively.
The $F$-pool is a set of $F$ values, e.g., $\{0.4, $ $ 0.5$ $, 0.6, $ $0.7, $ $0.8, $ $0.9\}$, and the $\CR$-pool is a set of the $\CR$ values, e.g.,  $\{0.1, $ $0.2, $ $0.3, $ $0.4, $ $0.5, $ $0.6, $ $0.7, $ $0.8, $ $0.9\}$.
At the beginning of the search, each individual $\vector{x}_{i,t}$ is randomly assigned values for $F_{i,t}$ and $\CR_{i,t}$ from each pool.
During the search, successful parameter sets are inherited by the individual in the next iteration.
Parameter sets that fail are reinitialized.

%% The overall generalized parameter adaptation method in EPSDE is described in Algorithm \ref{alg:general-epsde} in the supplemental material.

%% \subsection{JADE}
%% \label{sec:jade}

\noindent {\bf $\bullet$ PAM-JADE}:
PAM-JADE \cite{ZhangS09} uses two adaptive meta-parameters $\mu_{F} \in (0,1]$ and  $\mu_{\CR} \in [0,1]$ for parameter adaptation.
At the beginning of the search, $\mu_{F}$ and $\mu_{\CR}$ are both initialized to 0.5, and adapted during the search.
In each iteration $t$, $F_{i,t}$ and $\CR_{i,t}$ are generated according to the following equations: $F_{i,t} = {\rm randc}(\mu_{F}, 0.1)$ and $\CR_{i,t} = {\rm randn}(\mu_{\CR}, 0.1)$.
%% %
%% %
%% \begin{align}  
%% \label{eqn:JADE_f_gen}
%% F_{i,t} &= {\rm randc}(\mu_{F}, 0.1)
%% %
%% %
%% \CR_{i,t} &= {\rm randn}(\mu_{\CR}, 0.1)
%% \end{align}
%
${\rm randc}(\mu_{F}, \sigma)$ are values selected randomly from a Cauchy distribution with location parameter $\mu_{F}$ and scale parameter $\sigma$.
${\rm randn}(\mu_{\CR}, \sigma^2)$ are values selected randomly from a normal distribution with mean $\mu_{\CR}$ and variance $\sigma^2$.
When $F_{i,t}> 1$, $F_{i,t}$ is truncated to 1, and when $F_{i,t} \leq 0$, the new $F_{i,t}$ is repeatedly generated in order to generate a valid value.
In case a value for $\CR_{i,t}$ outside of $[0,1]$ is generated, it is replaced by the limit value (0 or 1) closest to the generated value.

In each iteration $t$, successful $F$ and $\CR$ parameter pairs are added respectively to sets $\vector{S}^{F,t}$ and $\vector{S}^{\CR,t}$. 
We will use $\vector{S}$ to refer to $\vector{S}^{F,t}$ or $\vector{S}^{\CR,t}$ wherever the ambiguity is irrelevant or resolved by context.
At the end of the iteration, $\mu_{F}$ and $\mu_{\CR}$ are updated as: $\mu_{F} = (1 - c) \: \mu_{F} + c \: {\rm mean}_L(\vector{S}^{F,t})$ and $  \mu_{\CR} = (1 - c) \: \mu_{\CR} + c \: {\rm mean}_A(\vector{S}^{\CR,t})$, 
%% %
%% %
%% \begin{align}  
%% \small
%% \label{eqn:update_mu_f}
%%   \mu_{F} &= (1 - c) \: \mu_{F} + c \: {\rm mean}_L(\vector{S}^{F})\\
%% \label{eqn:update_mu_cr}
%%   \mu_{\CR} &= (1 - c) \: \mu_{\CR} + c \: {\rm mean}_A(\vector{S}^{\CR})
%% \end{align}
%
where the meta-level control parameter $c \in [0,1]$ is a learning rate, 
${\rm mean}_A(\cdot)$ is an arithmetic mean, and ${\rm mean}_L(\cdot)$ is a Lehmer mean which is computed as: ${\rm mean}_L(\vector{S}) = \sum_{s \in \vector{S}} s^2 / \sum_{s \in \vector{S}} s$.
%% %
%% \begin{align}  
%% \small
%% \label{eqn:lehmer_mean}
%% {\rm mean}_L(\vector{S}) = \frac{\sum_{s \in \vector{S}} s^2}{\sum_{s \in \vector{S}} s}
%% \end{align}
%
%$\vector{S}$ refers to either $\vector{S}^{F}$ or $\vector{S}^{\CR}$.
%As the search progresses, $\mu_{F}$ and $\mu_{\CR}$ should gradually approach the appropriate values for the given problem. %This sounds like there is one set of appropriate values for a given problem, but this paper is about tracking target F/C values which vary.

%The overall generalized parameter adaptation method in JADE is described in Algorithm \ref{alg:general-jade} in the supplemental material.

%\subsection{MDE}

\noindent {\bf $\bullet$ PAM-MDE}:
A parameter adaptation method in MDE \cite{IslamDGRS12} is similar to PAM-JADE and uses the meta-parameters $\mu_{F}$ and  $\mu_{\CR}$ for parameter adaptation of $F$ and $\CR$,  respectively.
In each iteration $t$, $F_{i,t}$ and $\CR_{i,t}$ are generated as same with PAM-JADE  respectively.
At the end of each iteration, $\mu_{F}$ and $\mu_{\CR}$ are updated as: $\mu_{F} = (1 - c_F) \: \mu_{F} + c_F \: {\rm mean}_P(\vector{S}^{F,t})$ and $\mu_{\CR} = (1 - c_{\CR}) \: \mu_{\CR} + c_{\CR} \: {\rm mean}_P(\vector{S}^{\CR,t})$, 
%% %
%% \begin{align}  
%% \small
%% \label{eqn:mde_update_mu_f}
%% \mu_{F} &= (1 - c_F) \: \mu_{F} + c_F \: {\rm mean}_P(\vector{S}^{F})\\
%% \label{eqn:mde_update_mu_cr}
%% \mu_{\CR} &= (1 - c_{\CR}) \: \mu_{\CR} + c_{\CR} \: {\rm mean}_P(\vector{S}^{\CR})
%% \end{align}
%
where $c_F$ and $c_{\CR}$ are uniformly selected random real numbers from $(0.0, 0.2]$ and $(0.0, 0.1]$,  respectively.
In contrast to JADE, the learning rates $c_F$ and $c_{\CR}$ are randomly assigned in each iteration $t$.
${\rm mean}_P(\cdot)$ is a power mean: ${\rm mean}_P(\vector{S}) = \bigr( \frac{1}{|\vector{S}|} \sum_{s \in \vector{S}} s^{1.5} \bigl)^{\frac{1}{1.5}}$
%
%
%
%
%% \begin{align}  
%% \small
%% {\rm mean}_P(\vector{S}) = \bigr( \frac{1}{|\vector{S}|} \sum_{s \in \vector{S}} s^{n} \bigl)^{\frac{1}{n}}, n=1.5
%% \end{align}
%% %

%The overall generalized parameter adaptation method in MDE is described in Algorithm \ref{alg:general-mde} in the supplemental material.

%% \subsection{SHADE}

\noindent {\bf $\bullet$ PAM-SHADE}:
PAM-SHADE \cite{TanabeF13} uses historical memories  $\vector{M}^{F} $ and $ \vector{M}^{\CR}$ for parameter adaption of $F$ and $\CR$, where $\vector{M}^{F} = (M^{F}_1, ..., M^{F}_H)^{\rm T}$ and $\vector{M}^{\CR} = (M^{\CR}_1, ..., M^{\CR}_H)^{\rm T}$.
Here, $H$ is a memory size, and all elements in $\vector{M}^{F} $ and $ \vector{M}^{\CR}$ are initialized to 0.5.
In each iteration $t$,  $F_{i,t}$ and $\CR_{i,t}$ used by each individual $\vector{x}^{i,t}$ are generated by randomly selecting an index $r_{i,t}$ from $\{1, ..., H\}$, and then applying the following formulas: $F_{i,t} = {\rm randc}(M^{F}_{r_{i,t}}, 0.1)$ and $\CR_{i,t} = {\rm randn}(M^{\CR}_{r_{i,t}}, 0.1)$
%
%
%
%% \begin{align}
%% %\footnotesize
%% \label{eqn:SHADE_f_gen}
%% F_{i,t} &= {\rm randc}(M^{F}_{r_{i,t}}, 0.1)\\  
%% \label{eqn:SHADE_cr_gen}
%% \CR_{i,t} &= {\rm randn}(M^{\CR}_{r_{i,t}}, 0.1)
%% \end{align}
%
If the values generated for $F_i$ and $\CR_i$ are outside the range $[0, 1]$, they are adjusted/regenerated according to the procedure described above for PAM-JADE.

At the end of the iteration, the memory contents in $\vector{M}^{F}$ and $\vector{M}^{\CR}$ are updated using the Lehmer mean as follows: $M^{F}_k = {\rm mean}_L(\vector{S}^{F,t})$ and $M^{\CR}_k = {\rm mean}_L(\vector{S}^{\CR,t})$.
%
%% %
%% \begin{align}
%% %\footnotesize
%% \label{eqn:shade_update_f}
%% M^{F}_k &= {\rm mean}_L(\vector{S}^F)\\
%% \label{eqn:shade_update_cr}
%% M^{\CR}_k &= {\rm mean}_L(\vector{S}^{\CR})
%% \end{align}
%
%
An index $k \in \{1, ..., H\}$ determines the position in the memory to update.
At the beginning of the search, $k$ is initialized to 1. 
Here, $k$ is incremented whenever a new element is inserted into the history.
If $k > H$, $k$ is set to 1.

%% {\bf We emphasize that in this paper, PAM-jDE, PAM-EPSDE, PAM-JADE, PAM-MDE, and PAM-SHADE refer to the parameter adaptation mechanisms used by jDE, EPSDE, JADE, MDE, and SHADE, as described above. They do {\it not} refer to the entire adaptive DEs which use these PAMs. The terms  ``jDE'', ``EPSDE'', ``JADE'', ``MDE'', and ``SHADE'' refer to the complete, adaptive DEs.}

%\input{limitation.tex}

\section{TPAM Simulation Framework}
\label{sec:proposed_method}

% XXX no need for a very short introducton to this section
%In this section, we first  define the TPAM framework (Section \ref{sec:tpam_description}), and then define three target functions ($g$) in  Section  \ref{sec:tpam_tf} which are used in later experiments.
%Finally, we discuss the proposed TPAM framework in Section \ref{sec:tpam_discussion}.

%\subsection{TPAM}
%\label{sec:tpam_description}

As described in Section \ref{sec:ade}, for each iteration $t$, PAMs in adaptive DE assign $F_{i,t}$, $\CR_{i,t}$ to each individual $\vector{x}^{i,t}$ in $\vector{P}^t = \{\vector{x}^{1,t}, ..., \vector{x}^{N,t} \}$.
Then, each trial vector $\vector{u}^{i,t}$ is generated using a mutation strategy with $F_{i,t}$ and a crossover method with $\CR_{i,t}$.
Finally, at the end of iteration $t$, a set of successful parameters is used for parameter adaptation.
In summary, for parameter adaptation, PAMs iterate the following three procedures: (1) generating a control parameter set $\{F, \CR\}$, (2) deciding whether $\{F, \CR\}$ is successful or failed, and (3) doing something which influences future parameter generation step (e.g., updating some internal data structure).

A key observation is that in most PAMs for DE, including all of the PAMs reviewed in Section \ref{sec:ade}, steps (1)--(3) above only depend on whether each trial vector generation is a success or a failure, according to Definition \ref{def:success-failure}. They do {\em not} depend on the absolute objective function values of the trial vectors. {\em This means that analyzing PAM behavior does not require modeling the absolute objective function values of the trial vectors which are generated by the control parameter trajectory output by a PAM -- a  model of trial vector success/failure is sufficient}. This allows us to greatly simplify the modeling framework.

Thus, parameter adaptation of PAMs in adaptive DE can be simulated by using a surrogate model deciding the success or failure in the procedure (2), instead of the actual solution evaluation by the objective function.
In the proposed TPAM framework, this decision is made based on target parameters $\theta^{{\rm target}}_1, ..., \theta^{{\rm target}}_{t^{\rm max}}$ generated by a predefined target function $g$ which PAMs should track.
TPAM {\em only} evaluates the tracking performance of PAMs to the target parameters, independent from the variation operators used and test functions (e.g., the Sphere function) for benchmarking EAs.

%Maximum number of parameter samplings ()

%http://math.stackexchange.com/questions/326537/how-to-write-down-formally-number-of-occurences
%https://en.wikipedia.org/wiki/Iverson_bracket

\IncMargin{0.5em}
\begin{algorithm}[t]
\SetKwInOut{Input}{input}\SetKwInOut{Output}{output}
%\scriptsize
\footnotesize
%\small
%\SetAlgoLined
\SetSideCommentRight
%
%% \Input{PAM, $F$, $p_a$}
%% \Output{$s$}
%% \BlankLine
%\tcp{\  Initialization phase}
$S^{{\rm total}} \leftarrow 0$, $t \leftarrow 1$, initialize meta-parameters of a PAM\;
\While{$t < t^{\rm max}$}{
  $\theta^{{\rm target}}_t \leftarrow g(t)$\;
  \For{$i \in \{1, ..., N\}$}{
    Sampling $\theta_{i,t}$ using the PAM\;
  }
  \For{$i \in \{1, ..., N\}$}{
    \uIf{${\rm rand[0, 1]} \leq p_a (\theta_{i,t})$} {
      $s_{i,t} \leftarrow TRUE$, $S^{{\rm total}} \leftarrow S^{{\rm total}} + 1$\;
    }
    \Else{
      $s_{i,t} \leftarrow FALSE$\;
    }
  }
  Update the meta-parameters of the PAM using $\{s_{1,t}, ..., s_{N,t}\}$\;
  $t \leftarrow t + 1$\;
}
\Return $r^{\rm succ} = \frac{S^{{\rm total}}}{t^{\rm max} \times N}$\;
%$d_{\theta} = \Delta d_{\theta} / $\;
\caption{The proposed TPAM framework}
\label{alg:tpam}
\end{algorithm}\DecMargin{0.5em}

Algorithm \ref{alg:tpam} shows the TPAM framework.
The parameter $\theta$ represents one of the following three parameters: (i) $F$, (ii) $\CR$, (iii) a pair of $F$ and $\CR$.
At the beginning of the simulation, meta-parameters of a PAM are initialized. %(Algorithm \ref{alg:tpam}, line 1).
Then, the following procedures are repeated until reaching the maximum number of iterations $t^{\rm max}$.

The target parameter $\theta^{{\rm target}}_t$ in each iteration $t$ is given by the target function $g$ (Algorithm \ref{alg:tpam}, line 3), where $g$ is an arbitrarily defined function of $t$.
Three target functions used in our study will be described in Section \ref{sec:tpam_tf}.
It is worth noting that $g$ can also be defined as a function of the number of function evaluations.

The parameter $\theta_{i,t}$, $i \in \{1, ..., N\}$ is generated according to each PAM (Algorithm \ref{alg:tpam}, line $4 \sim 5$).
After all the parameters have been generated, they are probabilistically labeled as successful or failed (Algorithm \ref{alg:tpam}, line $6 \sim 10$).
In this paper, $\theta_{i,t}$ is treated as the successful parameter with an acceptance probability $p_a (\theta_{i,t}) \in [0, p_a^{\rm max}]$ defined as follow:
\begin{align}
\label{eqn:accept_prob}
p_a (\theta_{i,t}) = \max (- \alpha d_{i,t} + p_a^{\rm max}, 0)
\end{align}
where $d_{i,t} = |\theta_{i,t} - \theta^{{\rm target}}_t|$, and $d_{i,t}$ is the distance between $\theta_{i,t}$ and $\theta^{{\rm target}}_t$.
The two parameters $\alpha > 0$ and $p_a^{\rm max} \in [0,1]$ control the difficulty of the model of the TPAM simulation.
$\alpha$ adjusts a slope of probability in Eq. \eqref{eqn:accept_prob}, and $p_a^{\rm max}$ is the maximum probability of $p_a (\theta_{i,t})$ -- a larger $\alpha$ value and a smaller $p_a^{\rm max}$ value makes a simulation model difficult.
In Eq. \eqref{eqn:accept_prob}, the smaller the distance $d_{i,t}$, the acceptance probability $p_a (\theta_{i,t})$ is linearly increasing.
In fact, when $d_{i,t}=0$, $p_a (\theta_{i,t})$ takes the maximum probability $p_a^{\rm max}$.

% (i.e., $\theta_{i,t} = \theta^{{\rm target}}_t$)

At the end of each iteration $t$, the meta-parameters of the PAM are updated according to the binary decision of success or failure  (Algorithm \ref{alg:tpam}, line $11$).
A performance indicator in the proposed TPAM is the percentage of successful parameters ($r^{\rm succ} \in [0,1]$) in the simulation (Algorithm \ref{alg:tpam}, line $12$).
A higher $r^{\rm succ}$ represents that the PAM is able to track a given target parameters $\theta^{{\rm target}}_1, ..., \theta^{{\rm target}}_{t^{\rm max}}$, and thus its tracking performance is good.

\subsection{Target function $g$ for TPAM}
\label{sec:tpam_tf}

Target parameters $\theta^{{\rm target}}_1, ..., \theta^{{\rm target}}_{t^{\rm max}}$ in TPAM are given by a target function $g$.
Thus, the information of PAMs gained by the TPAM simulation significantly depends on which types of $g$ is used.
In this paper, we introduce the following three target functions ($g^{\rm lin}$, $g^{\rm sin}$, and $g^{\rm ran}$).
Below, $n_t \in (0, 1]$ is the number of sampling parameters until iteration $t$ divided by the maximum number of sampling $N \times t^{\rm max}$.
The range of $\theta$ and $\theta^{{\rm target}}$ were set to $[0.0, 1.0]$ and $[0.1, 0.9]$ respectively.

% $g^{\rm lin}$: linear
% $g^{\rm sin}$: sine funciton
% $g^{\rm ran}$: random walk

The linear function based target function $g^{\rm lin}$ is formulated as follows:
\begin{align}
\small
\label{eqn:tf_linear_up}
g^{\rm lin/inc} (n_t) &= 0.4 \: n_t + 0.5\\
\label{eqn:tf_linear_down}
g^{\rm lin/dec} (n_t) &= -0.4 \: n_t + 0.5
\end{align}
On $g^{\rm lin/inc}$, the target parameter $\theta^{{\rm target}}$ is linearly increasing from 0.5 to 0.9, and $\theta^{{\rm target}}$ is linearly decreasing  from 0.5 to 0.1 on $g^{\rm lin/dec}$.
In Eq. \eqref{eqn:tf_linear_up} and \eqref{eqn:tf_linear_down}, we set the slope value to 0.4 such that $\theta^{{\rm target}} \in [0.1, 0.9]$.
The function $g^{\rm lin}$ is the simplest target function for the TPAM simulation.
By applying PAMs to the TPAM simulation with $g^{\rm lin/inc}$ and $g^{\rm lin/dec}$, whether they are able to track the monotonically changing target parameters or not can be investigated.
Also, by comparing the results on the two linear functions, the hidden bias of parameter adaptation in PAMs can be found out.

%Also, the bias of parameter adaptation in PAMs can be found out.

We define the sinusoidal function based  target function $g^{\rm sin}$ as follows:
\begin{align}
\small
\label{eqn:tf_sin}
g^{\rm sin} (n_t) = 0.4 \: {\rm sin} (\omega \, n_t) + 0.5
\end{align}
where the amplitude value and the initial phase to 0.4 and 0.5 respectively.
The angular frequency $\omega > 0$ in Eq. \eqref{eqn:tf_sin} controls a change amount of the target parameter by one iteration.
A larger $\omega$ value makes a simulation model with $g^{\rm sin}$ difficult for PAMs to track the target parameters.
By applying PAMs to the TPAM simulation with $g^{\rm sin}$, the tracking performance of PAMs on the target parameter periodically changing can be analyzed.

Finally, the target function $g^{\rm ran}$ simulating the random walk is formulated as follows ($t \geq 2$):
\begin{align}
\small
\label{eqn:tf_rw}
g^{\rm ran} (n_t) = g^{\rm ran} (n_{t-1}) + s \, {\rm rand}[-1, 1]
\end{align}
where for $t = 1$, $g^{\rm ran} (n_1) = 0.5$.
${\rm rand}[-1, 1]$ returns a uniformly distributed random number in the range $[-1, 1]$.
The step size for the random walk $s \in (0, 1]$ adjusts the amount of the perturbation by one iteration.
That is, $s$ controls the difficulty of tracking the target parameters in the TPAM simulation with $g^{\rm ran}$.
When a target parameter generated according to Eq. \eqref{eqn:tf_rw} exceeds the boundary values $0.1$ or $0.9$, it is reflected as follows:
\begin{align}
\small
\label{eqn:tf_reflection}
g^{\rm ran} (n_t) = \begin{cases}
2 \times 0.9 - g^{\rm ran} (n_t) & \:  {\rm if} \: g^{\rm ran} (n_t) > 0.9 \\
2 \times 0.1 - g^{\rm ran} (n_t) & \:  {\rm if} \: g^{\rm ran} (n_t) < 0.1
  \end{cases}
\end{align}
In contrast to $g^{\rm sin}$ defined in Eq. \eqref{eqn:tf_sin}, the target parameters generated by $g^{\rm ran}$ irregularly change.
By applying PAMs to the TPAM simulation with $g^{\rm ran}$, the tracking performance of PAMs on the target parameter irregularly changing can be investigated.

\subsection{Discussion on TPAM}
\label{sec:tpam_discussion}

As discussed in Section  \ref{sec:introduction}, previous work on PAMs for adaptive DE have been limited to relatively shallow, qualitative discussions about search performance.
In contrast, comparing the $r^{\rm succ}$ values obtained using TPAM allows quantitative comparisons regarding the adaptive capability of PAMs, e.g., $r^{\rm succ}($PAM-JADE$)$ is $X\%$ higher than $r^{\rm succ}($PAM-jDE$)$, so 
$r^{\rm succ}($PAM-JADE$)$ is $X\%$ more successful than $r^{\rm succ}($PAM-jDE$)$ with respect to tracking target control parameter values, and therefore a ``better'' adaptive mechanism in that sense.

The selection/replacement policy in DE is deterministic \cite{StornP97,PriceSL05}.
A trial vector $\vector{u}^{i,t}$ which is more fit that its parent $\vector{x}^{i,t}$ (i.e., 
$ f(\vector{u}^{i,t}) \leq f(\vector{x}^{i,t})$) {\it always} replaces its parent,  $\vector{x}^{i,t+1} = \vector{u}^{i,t}$.
TPAM assumes and exploits this deterministic replacement policy. Thus, TPAM can not be directly applied to EAs with nondeterministic replacement policies such as GAs with the roulette wheel selection method.

It is important to keep in mind that TPAM is a simulation framework for evaluating the ability of a given PAM to track a given target function $g$ -- TPAM is {\it not} a benchmark function for adaptive DEs.
Thus, ``iterations'' and ``number of (parameter) samples'' refer only to the corresponding operations in Algorithm \ref{alg:tpam}, and  do {\it not} correspond 1-to-1 to corresponding/similar terms related to amount of search performed (number of individuals) in a complete DE algorithm.
The reason we execute parameter sampling for some number of iterations $/$ number of samples is to evaluate PAM behavior over a sufficiently large window of activity -- this does not correspond to any specific number of search steps executed by a DE with that given PAM.

\begin{figure}[t]
\small
\newcommand{\widthvar}{0.23}
  \begin{center} 
\includegraphics[width=\widthvar\textwidth]{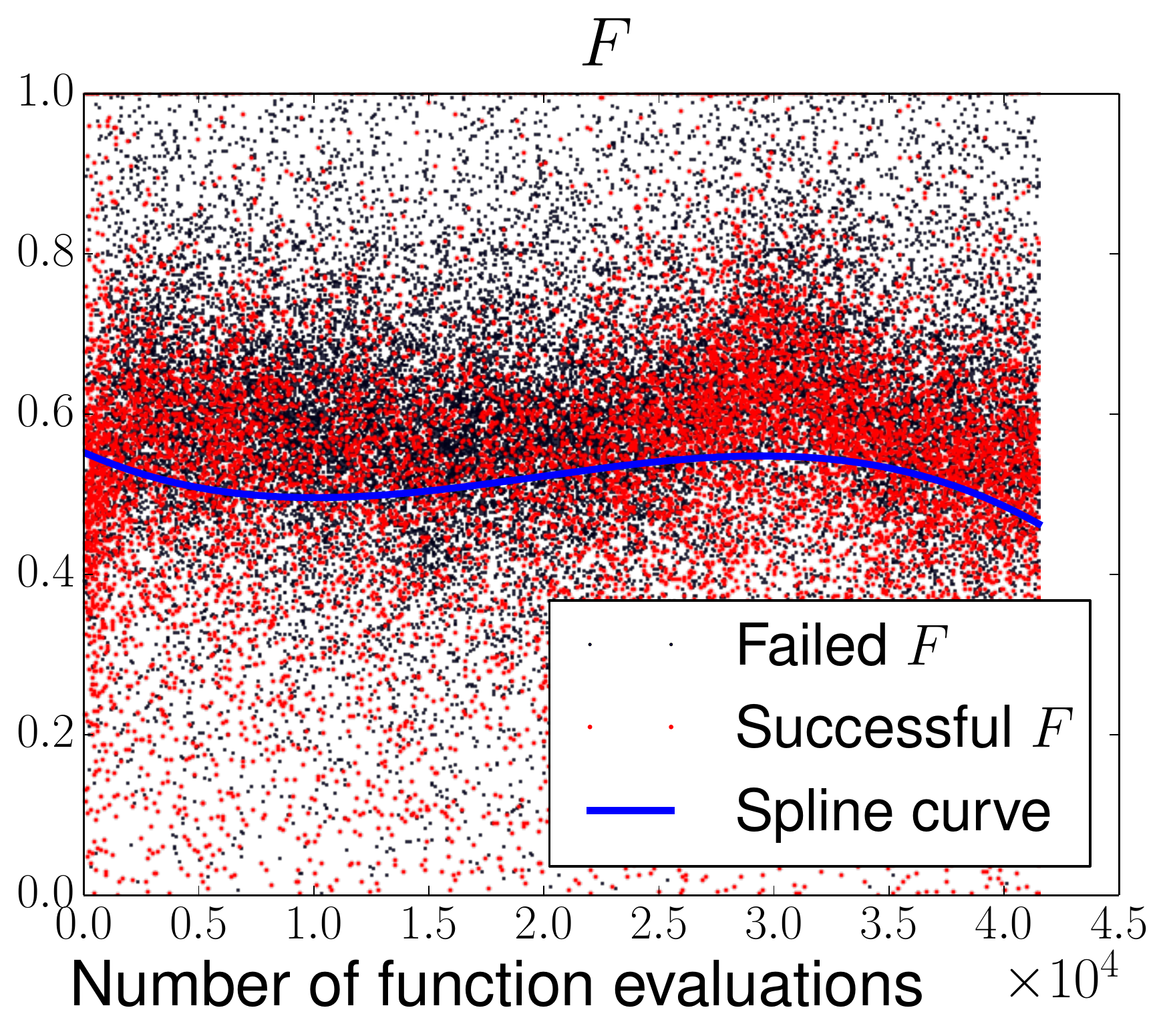}
\includegraphics[width=\widthvar\textwidth]{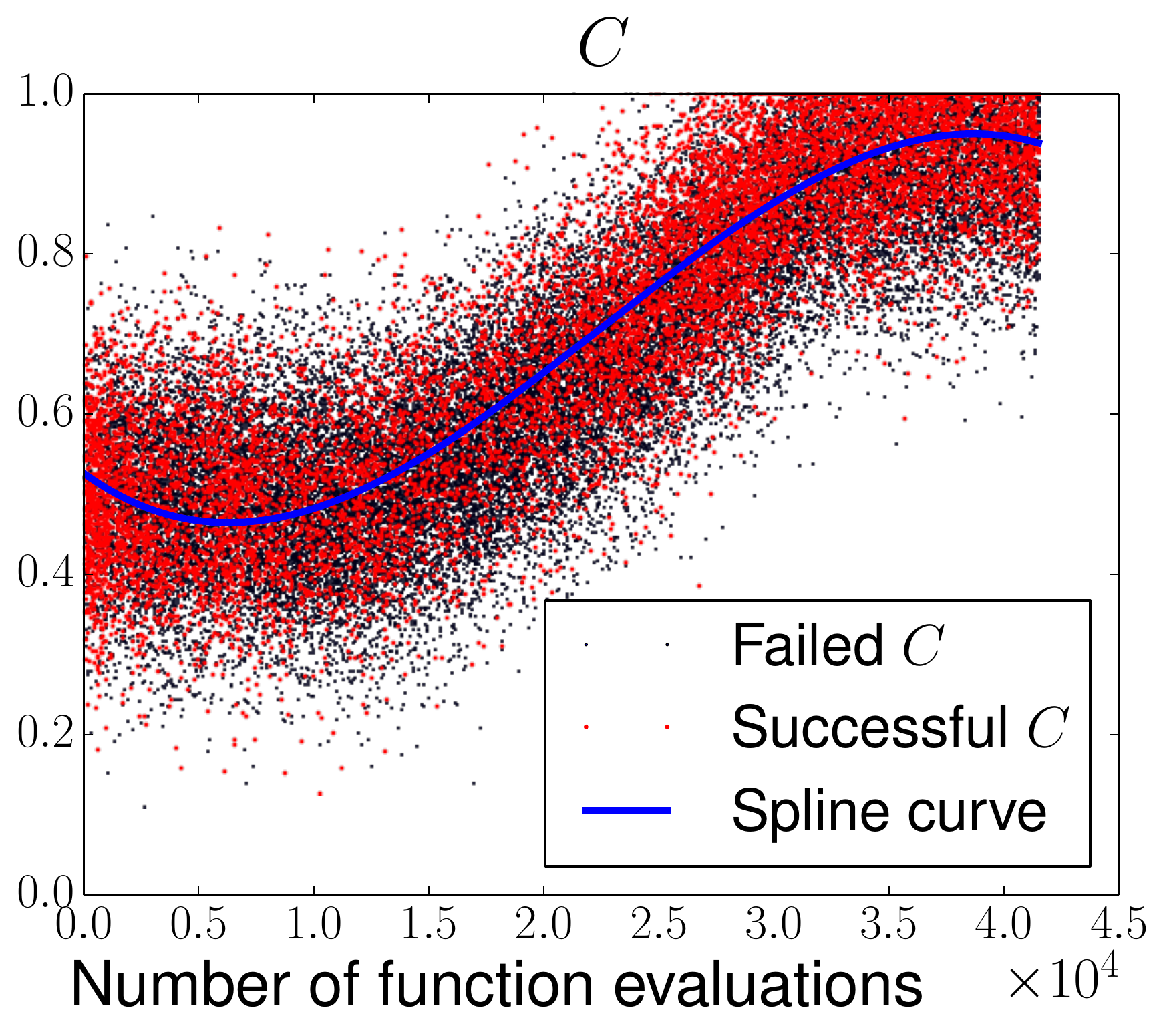}
\caption{
\small
All of the parameter values ($F$ and $\CR$) generated by an adaptive DE algorithm using the current-to-$p$best/1/bin and PAM-JADE on the 10-dimensional $f_8$ (Rosenbrock function) in the BBOB benchmarks.
The red and black points are successful and failed parameters respectively.
We also show smoothing spline curves for the successful parameter values.
Data from the median run is shown.
%Data from the median run out of 15 runs is shown.
%
%
}
\label{fig:success_failed_plot}
%%   \end{center}
%% \end{figure}
%
%% \begin{figure}[t]
%% \small
\renewcommand{\widthvar}{0.47}
%%   \begin{center} 
%% %
%
\label{fig:success_prob_f8}
\includegraphics[width=\widthvar\textwidth]{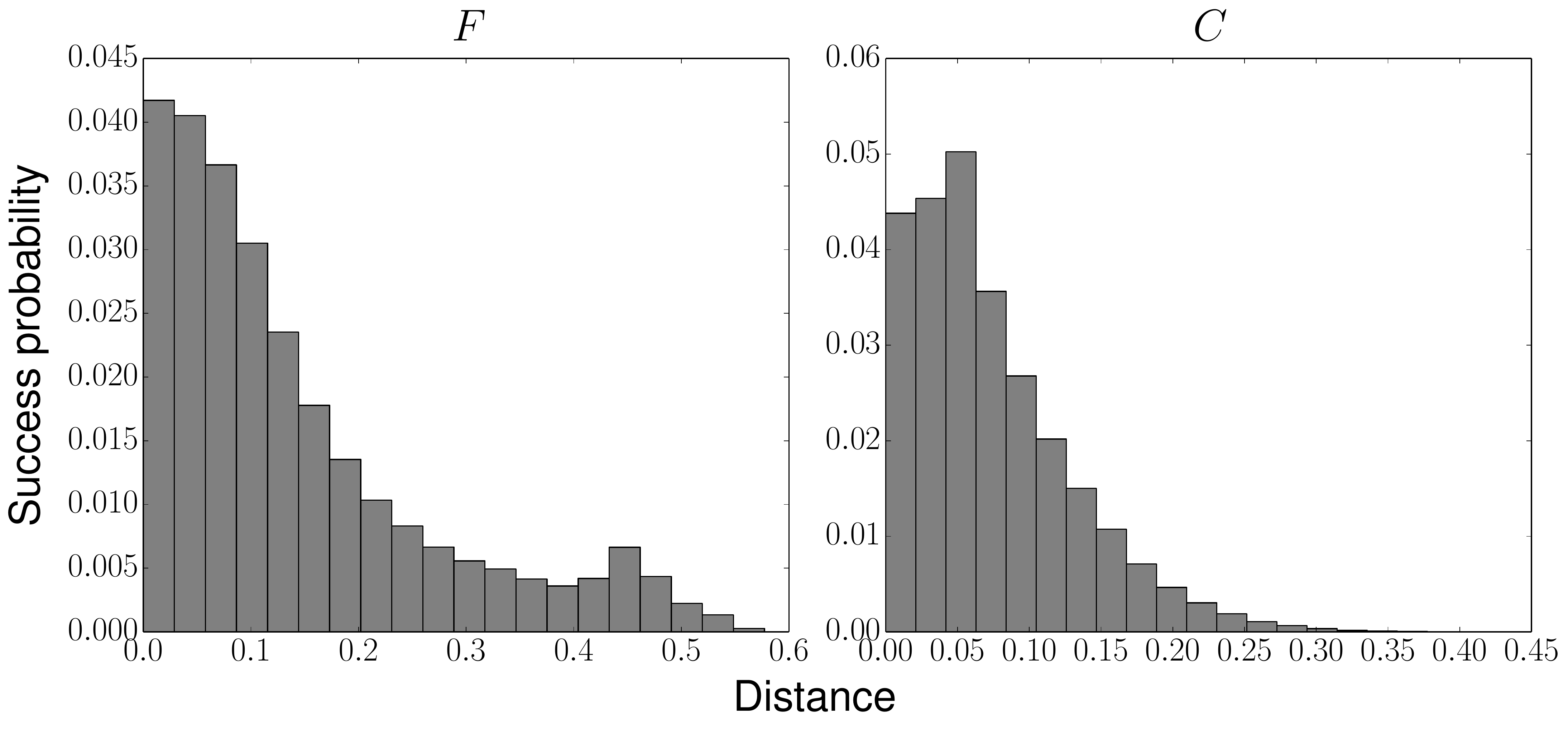}
\caption{
\small
Trial vector success probability as a function of the distance of the $F$ (left) and $\CR$ (right) parameters from  the spline curve on a run of an adaptive DE algorithm using the current-to-$p$best/1/bin and PAM-JADE on the 10-dimensional $f_8$ benchmark function.
Data from the all 15 run is shown.
}
\label{fig:success_prob}
  \end{center}
\end{figure}

According to Eq. \eqref{eqn:accept_prob}, 
the further $\theta$ is from $\theta^{{\rm target}}$, the lower its probability of success, $p_a (\theta)$.
This is intended to model the assumption that as $|\theta - \theta^{{\rm target}}|$ increases, $\theta$ becomes less and less appropriate for the current state of the search, and hence it becomes less likely for a trial vector generated using  $\theta$  to successfully replace its parent.
Below, we investigate the validity of this assumption.

Figure \ref{fig:success_failed_plot} shows all of the parameter values (including successful/unsuccessful values) generated by a run of an adaptive DE algorithm using the current-to-$p$best/1/bin and PAM-JADE (i.e., ``JADE'') on the 10-dimensional Rosenbrock function ($f_8$ in the BBOB benchmarks \cite{hansen2012fun}).
Figure \ref{fig:success_failed_plot} also shows smoothing spline curves for the successful parameter values.
PAM-JADE generates $F$ and $\CR$ values based on random numbers from a Cauchy distribution and a normal distribution, respectively (see Section \ref{sec:ade}), and so a diverse set of parameters is generated.
It can be seen that $F$ and $\CR$ values closer to the spline curves tend to result in more successes.

To verify that control parameter values closer to  the spline curve tend to result in more successful trial vectors,
Figure \ref{fig:success_prob} shows the trial vector success probability as a function of the  distance 
of the $F$ (left) and $\CR$ (right) parameters from 
the spline curve on a run of the adaptive DE using the current-to-$p$best/1/bin and PAM-JADE on the 10-dimensional $f_8$ benchmark function.  %{\bf XXX same run as Figure 1??}
For both $F$, and $\CR$, it can be seen that the success probability tends to drop monotonically as the distance from their respective  spline curves increases.
The above experiments shows that the assumption that 
the probability of generating successful trial vectors is highly correlated with the ability to generate control parameters values $\theta$ which accurately track a target parameter is justifiable.
Thus, Eq. \eqref{eqn:accept_prob} is a reasonable model for the success probability $p_a (\theta_{i,t})$.
Although the linear function is used in Eq. \eqref{eqn:accept_prob} in this paper, future work will investigate other types of functions (e.g., the gamma distribution function).

%\section{Evaluating Parameter Adaptation Methods Using TPAM}
\section{Evaluating PAMs Using TPAM}
\label{sec:experiment}

\subsection{Experimental settings}
\label{sec:experimental-setting}

We investigate the tracking performance of PAMs of widely used five adaptive DEs (PAM-jDE, PAM-JADE, PAM-EPSDE, PAM-MDE, and PAM-SHADE) described in Section \ref{sec:ade} on the TPAM framework.

%For PAM$^{\rm JADE}$ and PAM-SHADE, $c$ and $H$ were set to $0.1$ and $10$ respectively.

The population size $N$ was set to 50.
The maximum number of iterations $t^{\rm max}$ was $1 \, 000$.
The 101 independent runs were performed.
For each PAM, we used the control parameter value suggested in the original papers as follows: $\tau_F  = \tau_{\CR} = 0.1$ for PAM-jDE, $c=0.1$ for PAM-JADE, and $H=10$ for PAM-SHADE.
In the original implementation, PAM-jDE generates the $F$ values in the range $[0.1, 1]$ as described in Eq. \eqref{eqn:jde-f}, but for a fair comparison we modified the range to $[0, 1]$.
For the same reason with PAM-jDE, we set $F$-pool $=\{0.0,0.1, ..., 0.9, 1.0 \}$ and $\CR$-pool $=\{0.0,0.1, ..., 0.9, 1.0 \}$ for PAM-EPSDE.
These modifications allow PAM-jDE and PAM-EPSDE to generate the $F$ and $\CR$ values in the range $[0,1]$.
Also, for a fair comparison, the initial $F_{i,t}$ and $\CR_{i,t}$ values for PAM-jDE and PAM-EPSDE were set to 0.5 as with the initial values of the meta-parameters of PAM-JADE, PAM-MDE, and PAM-SHADE.

%In our preliminary experiments,

$\alpha$ and $p_a^{\rm max}$ in Eq. \eqref{eqn:accept_prob} are the two control parameters for the proposed TPAM.
In our preliminary experiments, we confirmed that the $r^{\rm succ}$ values of all the five PAMs are monotonically decreasing when $\alpha$ increasing.
Due to space constraints, we show only the results of the TPAM simulation with $\alpha = 1$.
On the other hand, we used the value of $p_a^{\rm max} \in \{0.1, 0.2, ..., 1\}$.

%set the $\alpha$ value to $0.5, 1$, and $3$ and 

\begin{figure}[t]
\small
\newcommand{\widthvar}{0.46}
  \begin{center} 
\includegraphics[width=\widthvar\textwidth]{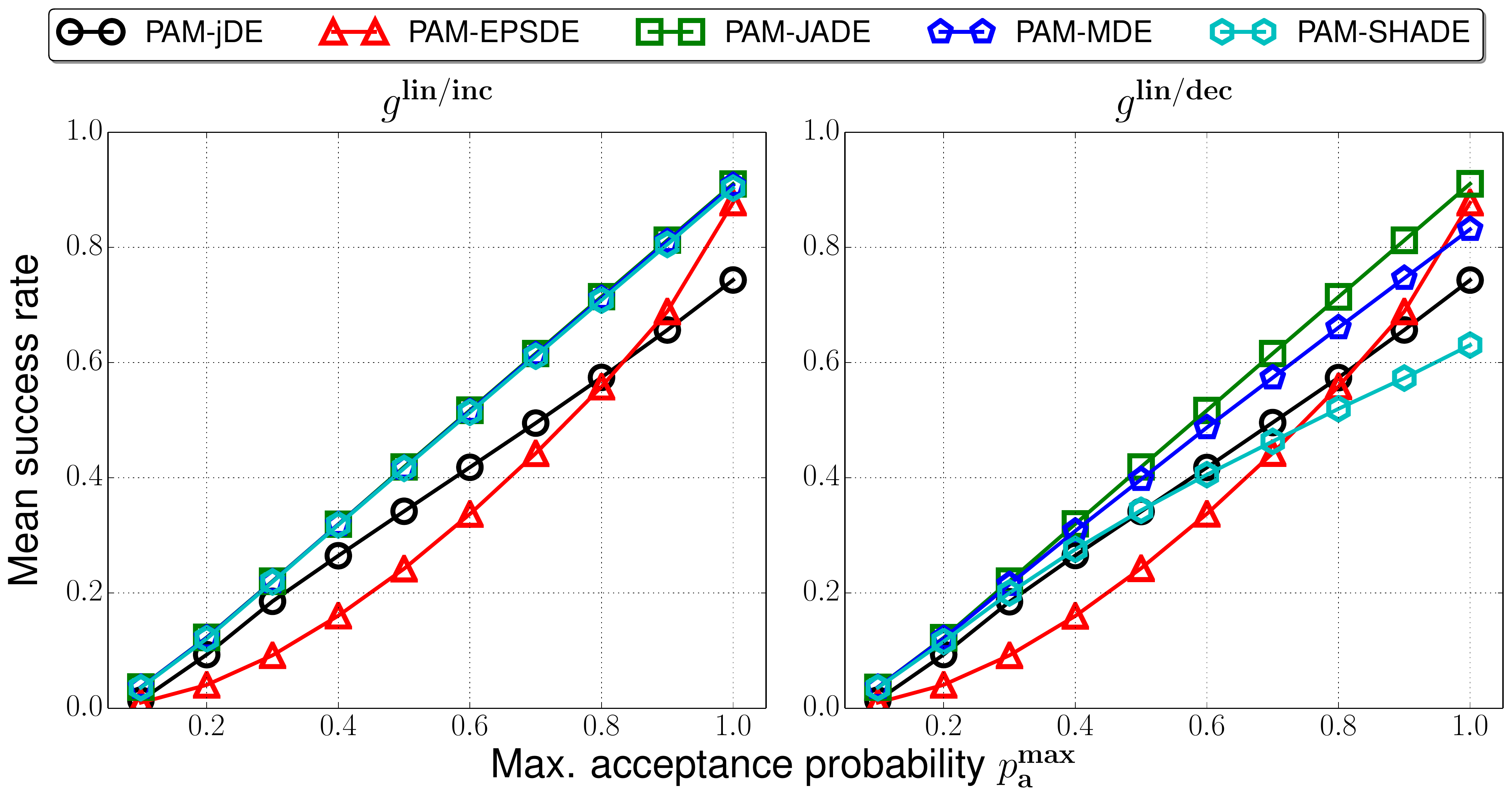}
\caption{
\small
TPAM simulation results for Target function $g^{\rm lin/inc}$ and $g^{\rm lin/dec}$.
The x-axis shows $p_a^{\rm max}$ in Eq. \ref{eqn:accept_prob}, and the y-axis shows the mean $r^{\rm succ}$ for 101 runs (higher is better).
%Error bars show standard deviation.
%
}
\label{fig:target_func_linear}
  \end{center}
\end{figure}

\begin{figure*}[t]
\small
\newcommand{\widthvar}{0.91}
  \begin{center} 
\includegraphics[width=\widthvar\textwidth]{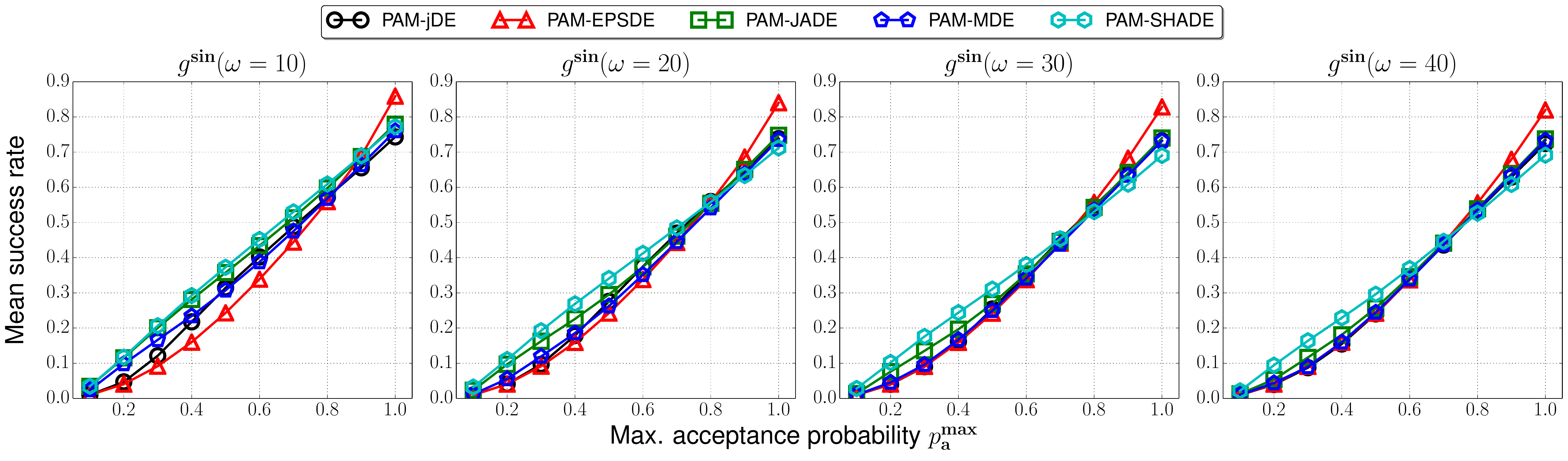}
\caption{
\small
TPAM simulation results for Target function $g^{\rm sin}$ for $\omega \in \{10, 20, 30, 40\}$.
The x-axis shows $p_a^{\rm max}$ in Eq. \ref{eqn:accept_prob}, and the y-axis shows the mean $r^{\rm succ}$ for 101 runs (higher is better).
%Error bars show standard deviation.
}
\label{fig:target_func_sin}
%%   \end{center}
%% \end{figure*}
%
%%  \begin{figure*}[t]
%% \small
%% \newcommand{\widthvar}{0.95}
%%  \begin{center} 
%
\includegraphics[width=\widthvar\textwidth]{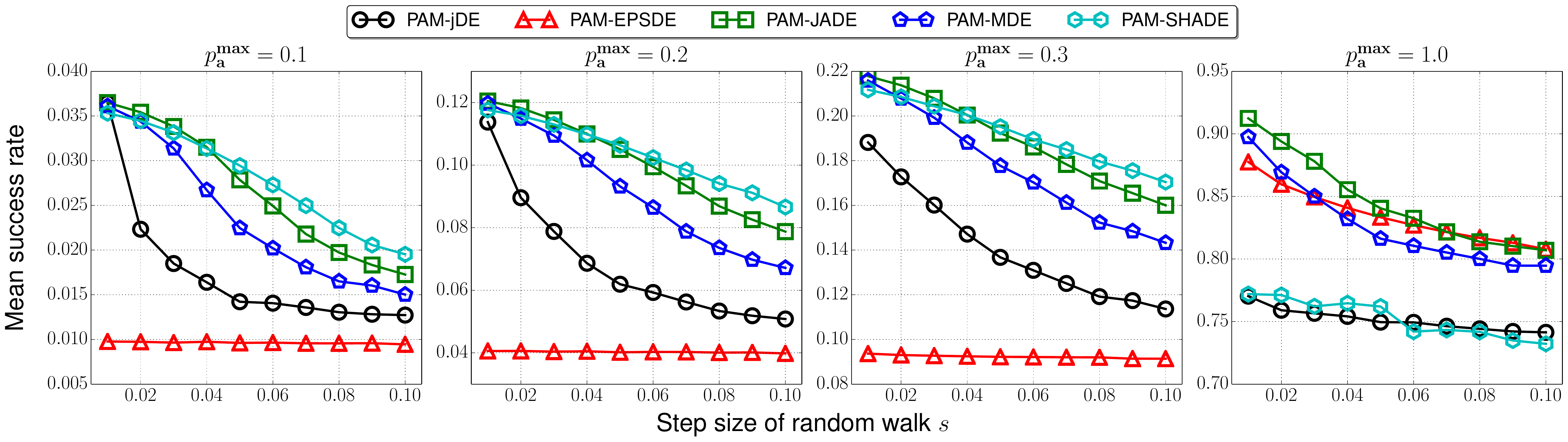}
\caption{
\small
TPAM simulation results for Target function $g^{\rm ran}$ for $p_a^{\rm max} \in \{0.1, 0.2, 0.3, 1\}$.
The x-axis shows the step size $s$ of random walk, and the y-axis shows the mean $r^{\rm succ}$ for 101 runs (higher is better).
%Error bars show standard deviation.
}
\label{fig:target_func_rw}
  \end{center}
\end{figure*}

\subsection{Tracking results for each target function}
\label{sec:results_each_tf}

Here, we evaluate the tracking performance of the five PAMs on the TPAM simulation with the three target functions $g^{\rm lin}$, $g^{\rm sin}$, and $g^{\rm ran}$.
We investigated the three types of parameters (i) $\CR$, (ii) $F$, and (iii) a pair of $F$ and $\CR$, but their qualitative tendency is not significantly different each other, and so we provide only the results of (i) $\CR$. %, but the others can be found in Figure XXX $\sim$ XXX in the supplemental file.
%Due to space constraints, we provide only the results of (i) $\CR$. %, but the others can be found in Figure XXX $\sim$ XXX in the supplemental file.

\subsubsection{Results on $g^{\rm lin}$}
\label{sec:results_tf_linear}

Figure \ref{fig:target_func_linear} shows the results of running the TPAM simulations on the target functions $g^{\rm lin/inc}$ (Eq. \eqref{eqn:tf_linear_up}) and  $g^{\rm lin/dec}$  (Eq. \eqref{eqn:tf_linear_down}).
The target functions  $g^{\rm lin/inc}$ and $g^{\rm lin/dec}$ simply linearly increase/decrease the target parameter $\theta^{{\rm target}}$, respectively. 
 There is very little difference among the success rates of PAM-jDE, PAM-EPSDE, and PAM-JADE on $g^{\rm lin/inc}$ and $g^{\rm lin/dec}$.
In contrast, PAM-MDE and PAM-SHADE tend to have a lower success rate on $g^{\rm lin/dec}$ compared to $g^{\rm lin/inc}$ for $p_a^{\rm max} \geq 0.3$.
In particular, PAM-SHADE has the worst tracking performance among all PAMs for $p_a^{\rm max} \in [0.8, 1.0]$ on $g^{\rm lin/dec}$.
A speculative explanation for this is that 
for parameter updates, PAM-MDE and PAM-SHADE use the power mean and Lehmer mean respectively, which tend to be pulled up to higher values, unlike arithmetic means.
Thus, on $g^{\rm lin/dec}$, where the target parameter $\theta^{{\rm target}}$ monotonically decreases,  PAM-MDE and PAM-SHADE have difficulty tracking the target, resulting in relatively low success rates compared to $g^{\rm lin/inc}$.

For both $g^{\rm lin/inc}$ and $g^{\rm lin/dec}$, $r^{\rm succ}$ tends to increase monotonically for all PAMs as $p_a^{\rm max}$ increases from 0 to 1. 
At $p_a^{\rm max} = 0.1$, all PAMs have almost the same $r^{\rm succ}$.
As  $p_a^{\rm max}$ increases, the relative success rate of PAM-jDE compared to other PAMs decreases.
PAM-jDE reinitializes the parameter values in the range $[0,1]$ with some certain probability (see Section \ref{sec:ade}), and as a result, the increase of its success rate is not as large as the remaining PAMs.
% ({\bf any possible explanation why?}).
The relative success rate of PAM-EPSDE increases as $p_a^{\rm max}$ increases, likely because PAM-EPSDE continues to use the same parameter value as long as it keeps succeeding, which is a good fit for $p_a^{\rm max} = 1$.
In contrast, for low maximum acceptance probabilities such as 
 $p_a^{\rm max} = 0.1$,  PAM-JADE, PAM-MDE, and PAM-SHADE had the highest average success rate.
This is likely because PAM-JADE, PAM-MDE, and PAM-SHADE generate parameter values which are close to values which have recently succeeded, so even if $p_a^{\rm max}$ is low, these approaches allocate a significant fraction of their samples around the target parameter values.

\subsubsection{Results on $g^{\rm sin}$}
\label{sec:results_tf_sin}

Figure \ref{fig:target_func_sin} shows the results of running TPAM using the target function $g^{\rm sin}$ (Eq. \eqref{eqn:tf_sin}), for
 the angular frequency $\omega \in \{10, 20, 30, 40 \}$.
For all PAMs, $r^{\rm succ}$ decreases as $\omega$ increases.
This is because as $\omega$ increases, the target parameter value changes more rapidly, making it more difficult for the PAMs to track the target parameters.

For all values of $\omega$, PAM-EPSDE achieves the highest $r^{\rm succ}$ for  $p_a^{\rm max} \in [0.9,1]$.
However, as $p_a^{\rm max}$ decreases, PAM-EPSDE performs worse than other PAMs, most likely due to the same reason as discussed in Section \ref{sec:results_tf_linear}.

PAM-SHADE has the worst performance among the PAMs for $\omega \geq 20$ and $p_a^{\rm max} \in [0.8,1.0]$.
However, the lower the value of  $p_a^{\rm max}$, the better PAM-SHADE performs compared to other PAMs, and this trend
strengthens as $\omega$ (the rate of change of the target parameter value) increases -- in particular, note that the difference between $r^{\rm succ}($PAM-SHADE$)$ and $r^{\rm succ}($PAM-JADE$)$  increases with $\omega$.
In other words, the more difficult it is to follow the target value, the better PAM-SHADE performs compared to other PAMs.

\subsubsection{Results on $g^{\rm ran}$}
\label{sec:results_tf_random_walk}

 Figure \ref{fig:target_func_rw} shows the results of running TPAM using the target function $g^{\rm ran}$ (Eq. \eqref{eqn:tf_rw}). As the step size $s$ increases, the rate of the random walk increases, making it increasingly more difficult for a PAM to track the target parameter. Due to space constraints, we only show results for $p_a^{\rm max} \in \{0.1, 0.2, 0.3, 1 \}$.

%Figure \ref{fig:target_func_rw}\subref{fig:rw_beta1.0} % XXX the trend is visible for all beta, not only beta~1.0.

Figure \ref{fig:target_func_rw}
shows that as $s$ increases, $r^{\rm succ}$ tends to decrease monotonically for all PAMs.
Thus, similar to our observations for $g^{\rm sin}$ above, $r^{\rm succ}$ tends to fall for all PAMs on $g^{\rm ran}$as the rate of change of the target parameter increases.
PAM-EPSDE performs well when $p_a^{\rm max} = 1$, as it did for $g^{\rm lin}$ and $g^{\rm sin}$.
However, Figure \ref{fig:target_func_rw} shows that for $p_a^{\rm max} \in \{0.1, 0.2, 0.3\}$, PAM-EPSDE has the worst tracking performance.
PAM-JADE has the best tracking performance among all PAMs when $s$ is small ($< 0.05$). 
However, for larger values of $s$, i.e., for rapid random walks, PAM-SHADE outperforms PAM-JADE.
The tracking performance of PAM-MDE was dominated by PAM-JADE for all $p_a^{\rm max}$ and $s$.

%{\bf XXX: The line labels in Figures 5a-d should be moved -- right now, they're covering up the results for small $s$}

%% %{fig:rw_beta1.0}

\subsection{Detailed analysis of target tracking behavior by each PAM}
\label{sec:metaparameters_on_tpam}

\begin{figure}[htp]
\small
\newcommand{\widthvar}{0.22}
  \begin{center} 
\subfloat[$g^{\rm lin/dec}$]{
\label{subfig:metaparams_linear0.1}
\includegraphics[width=\widthvar\textwidth]{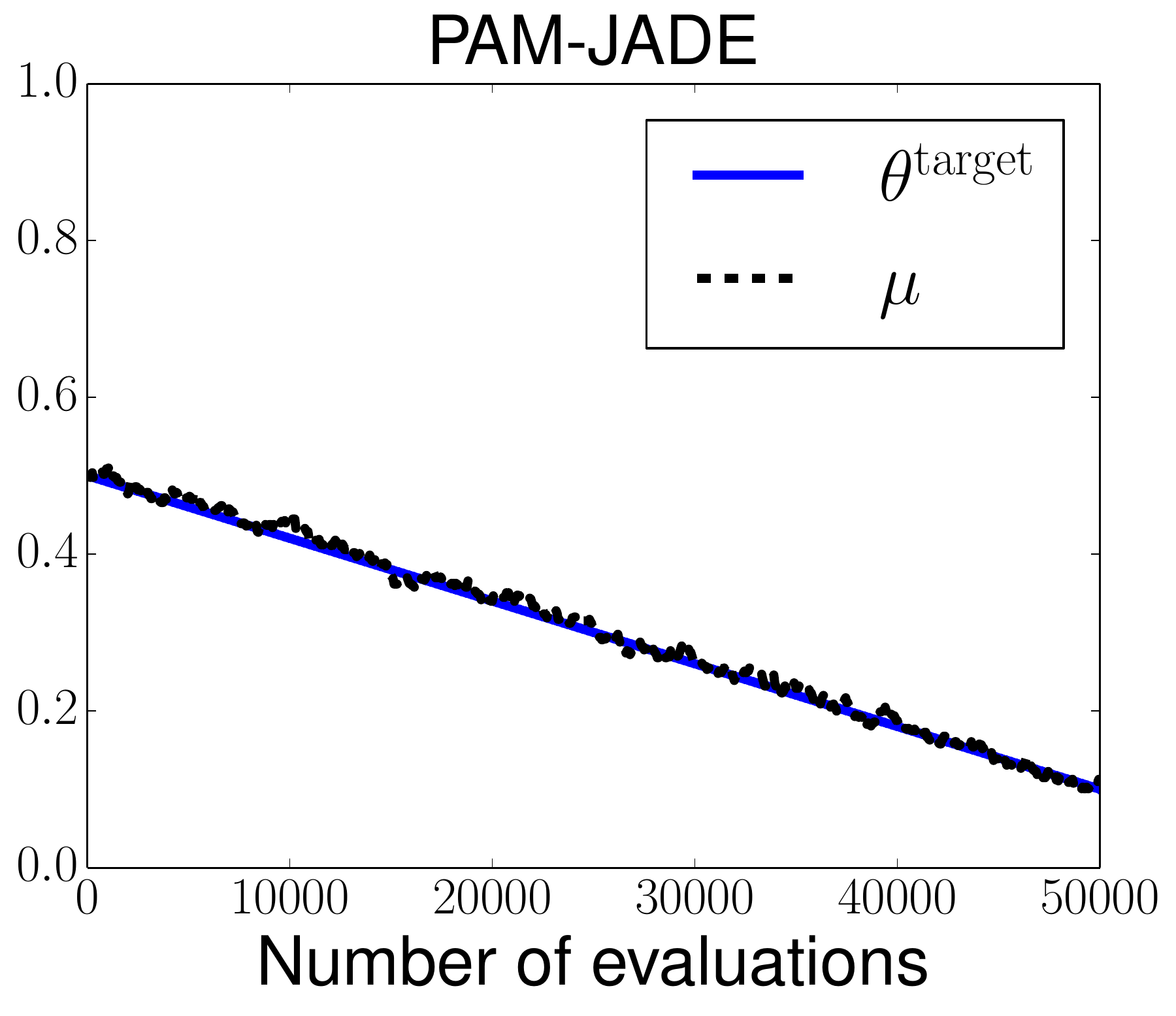}
\includegraphics[width=\widthvar\textwidth]{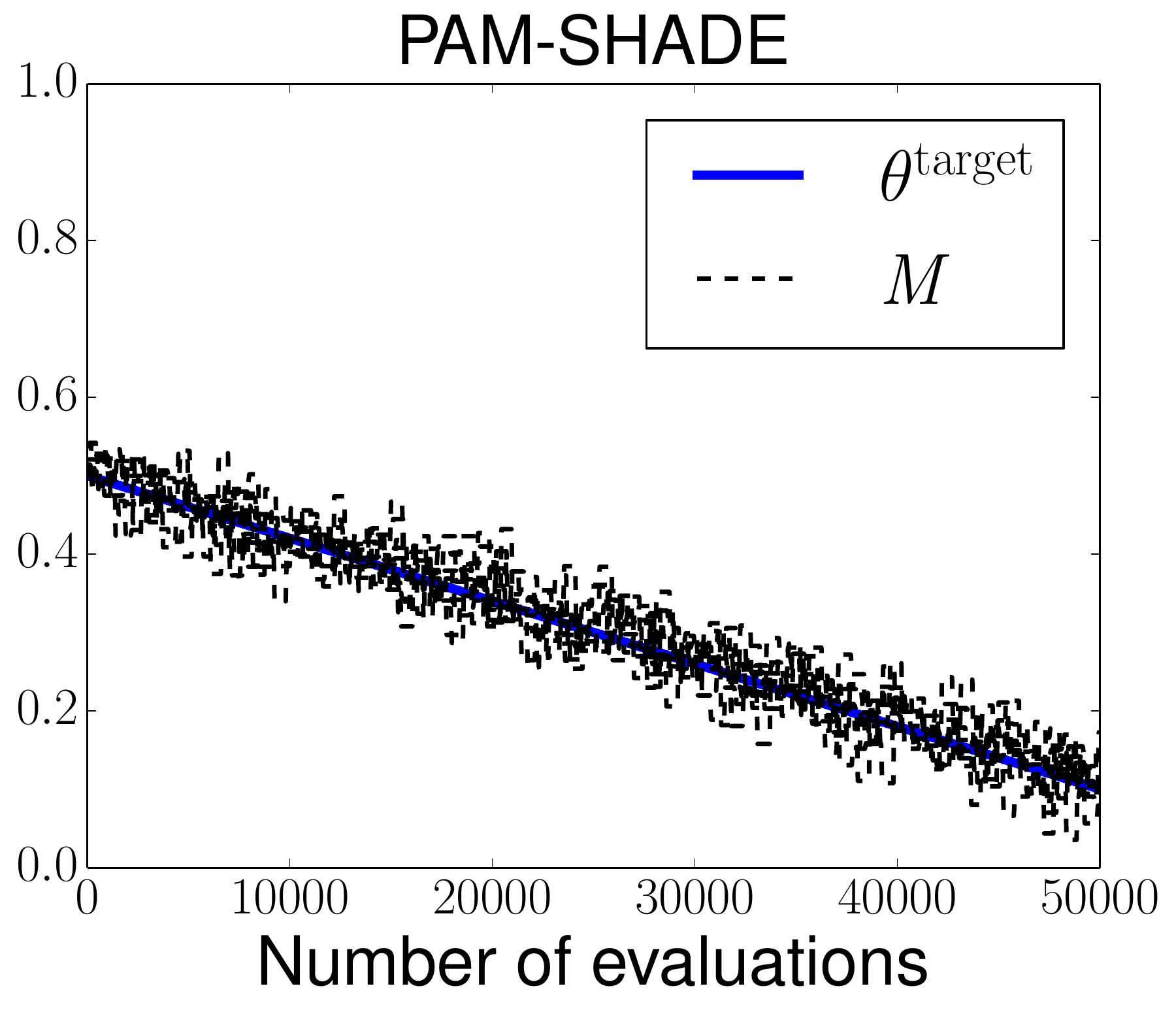}
}
\\[-0.1em]
\subfloat[$g^{\rm sin}$ $(\omega = 10)$]{
\label{subfig:metaparams_omega10}
\includegraphics[width=\widthvar\textwidth]{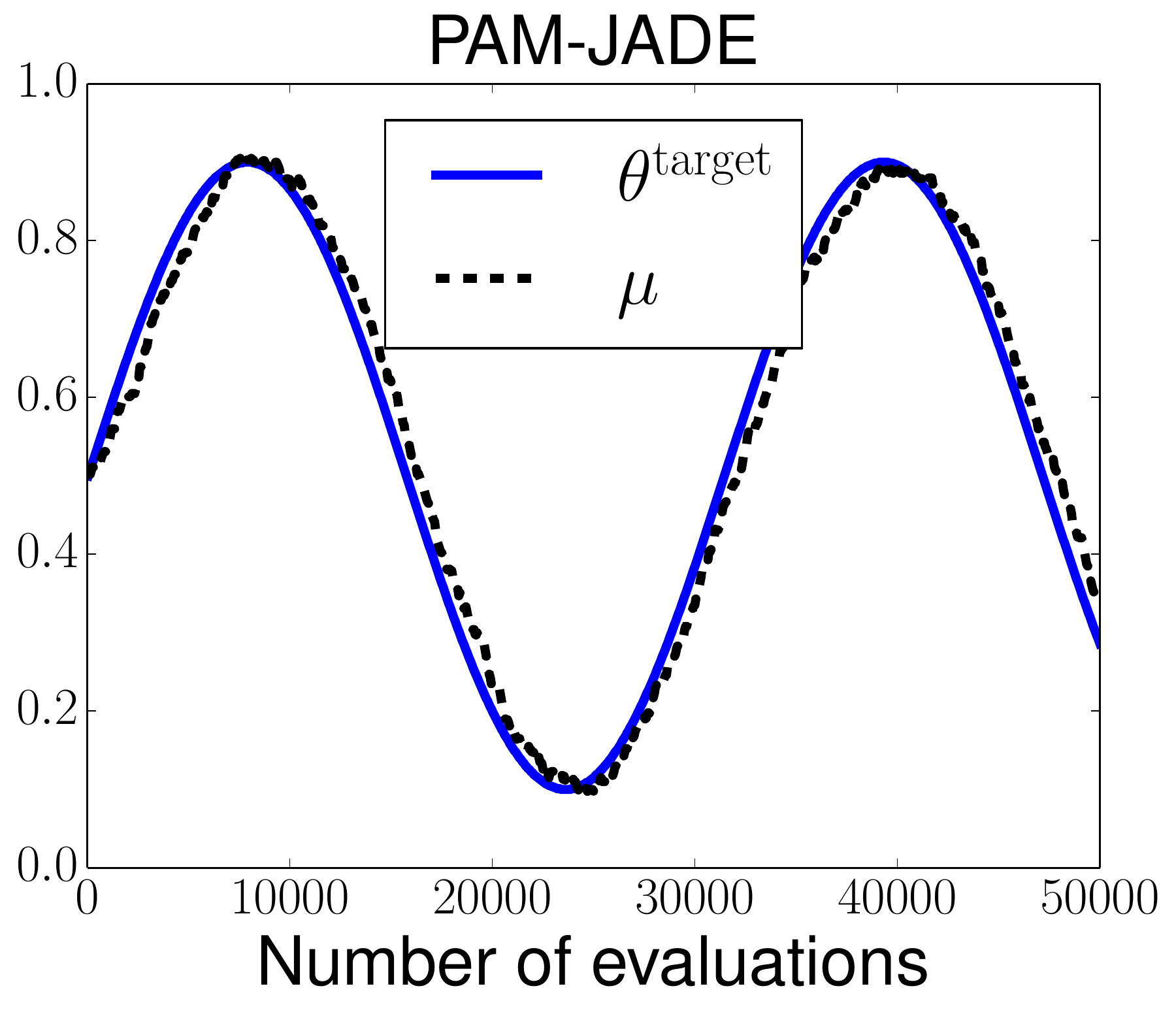}
\includegraphics[width=\widthvar\textwidth]{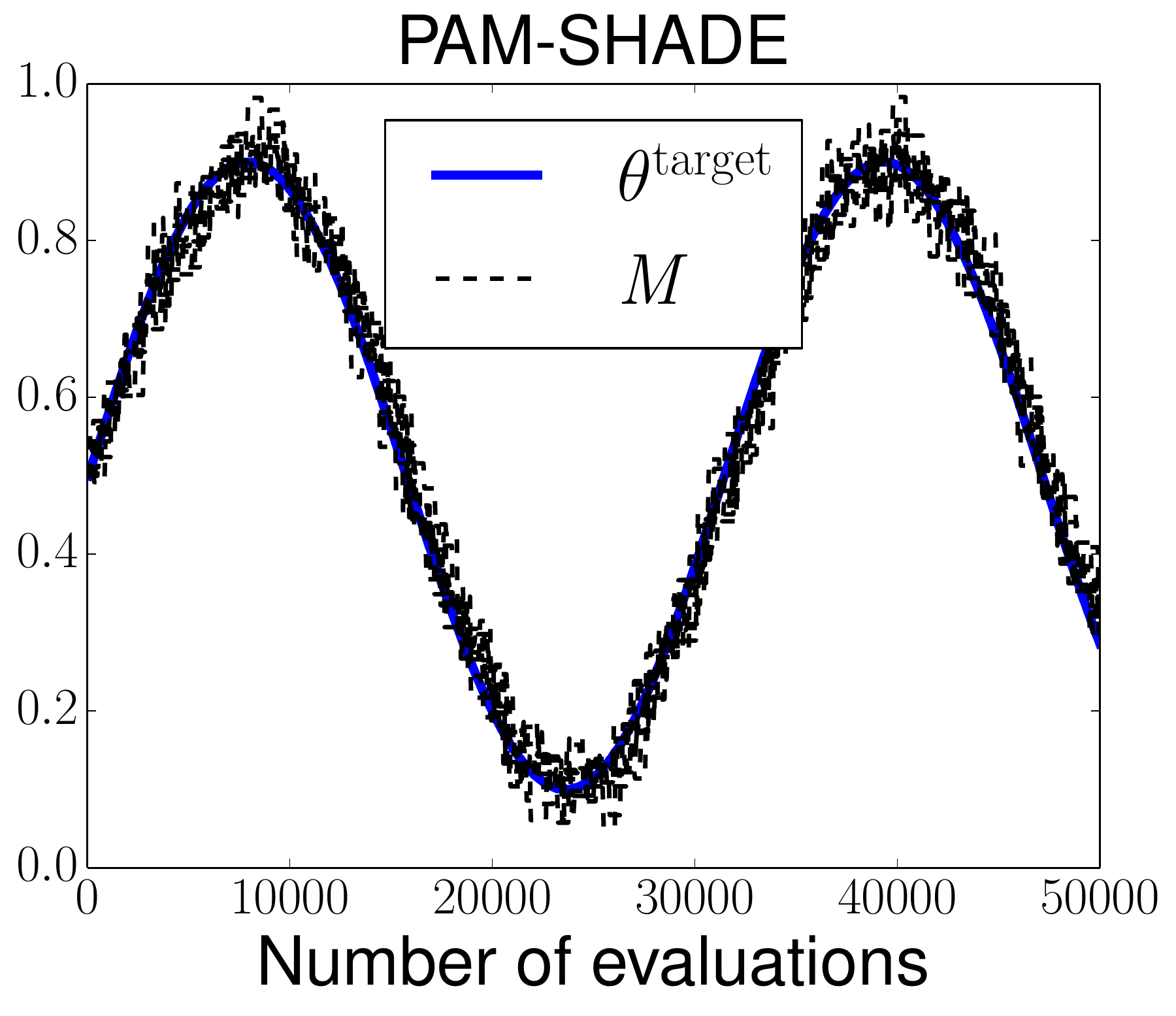}
}
\\[-0.1em]
\subfloat[$g^{\rm sin}$ $(\omega = 40)$]{
\label{subfig:metaparams_omega40}
\includegraphics[width=\widthvar\textwidth]{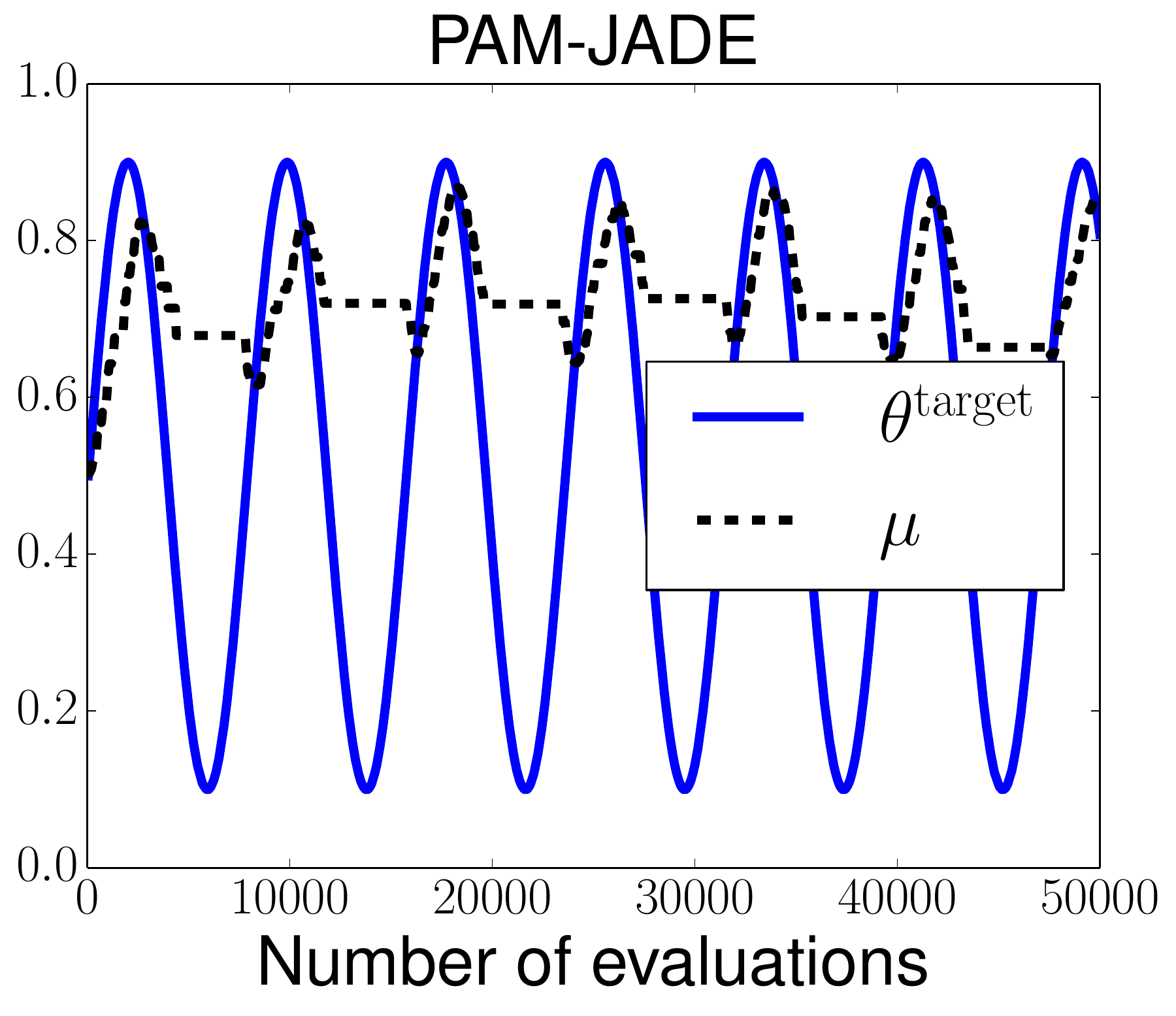}
\includegraphics[width=\widthvar\textwidth]{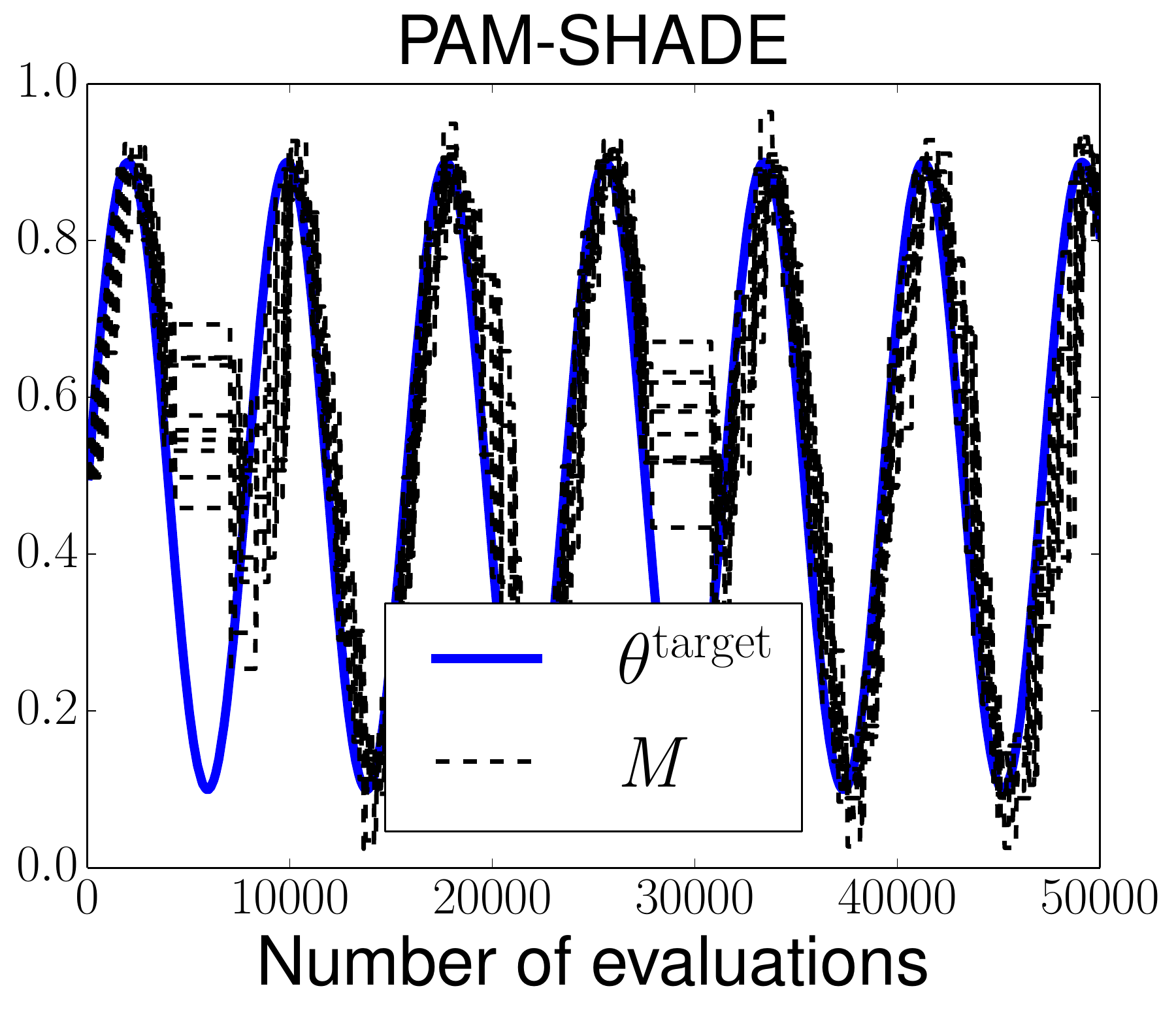}
}
\\[-0.1em]
\subfloat[$g^{\rm ran}$ $(s = 0.04)$]{
\label{subfig:metaparams_s0.04}
\includegraphics[width=\widthvar\textwidth]{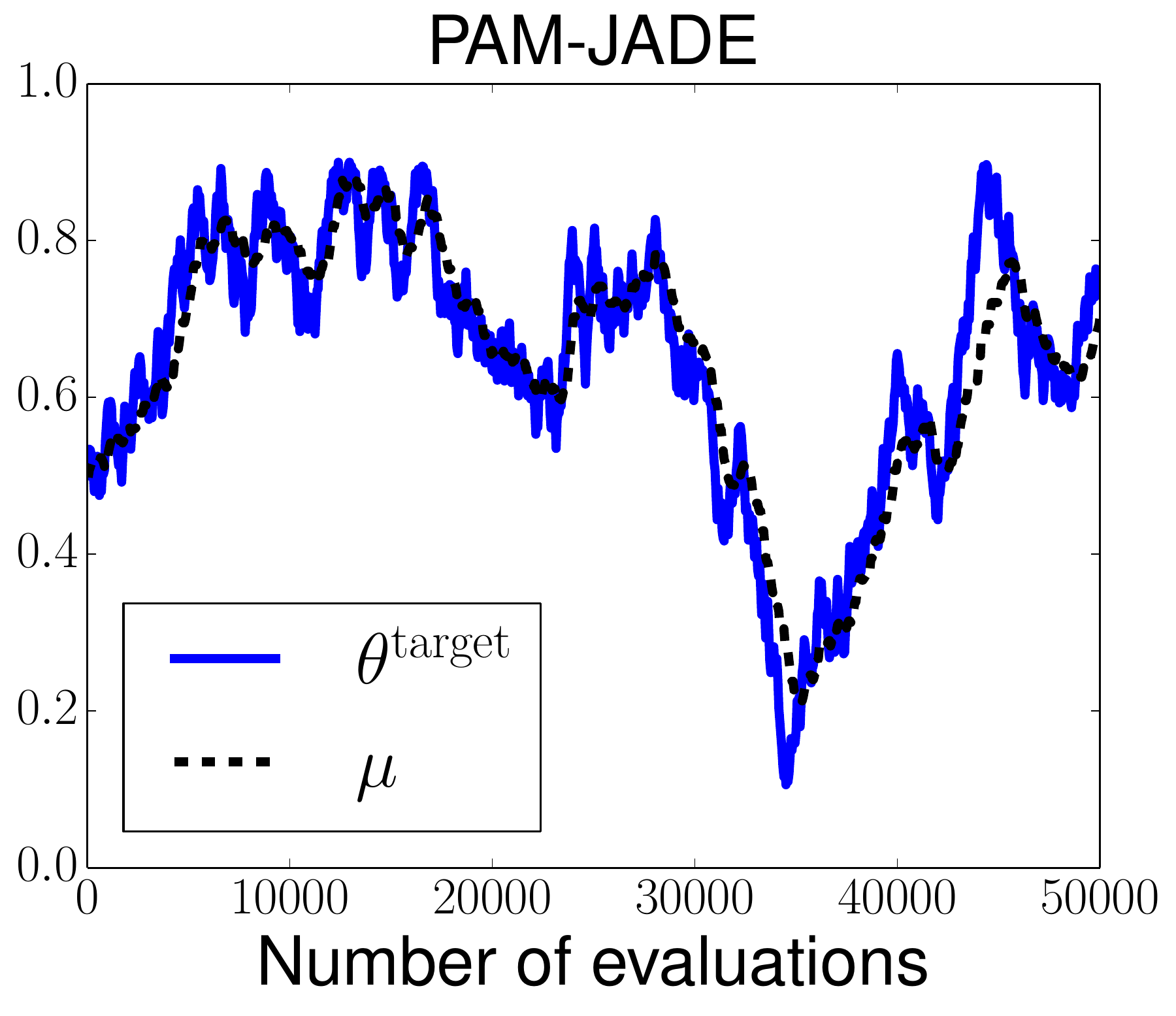}
\includegraphics[width=\widthvar\textwidth]{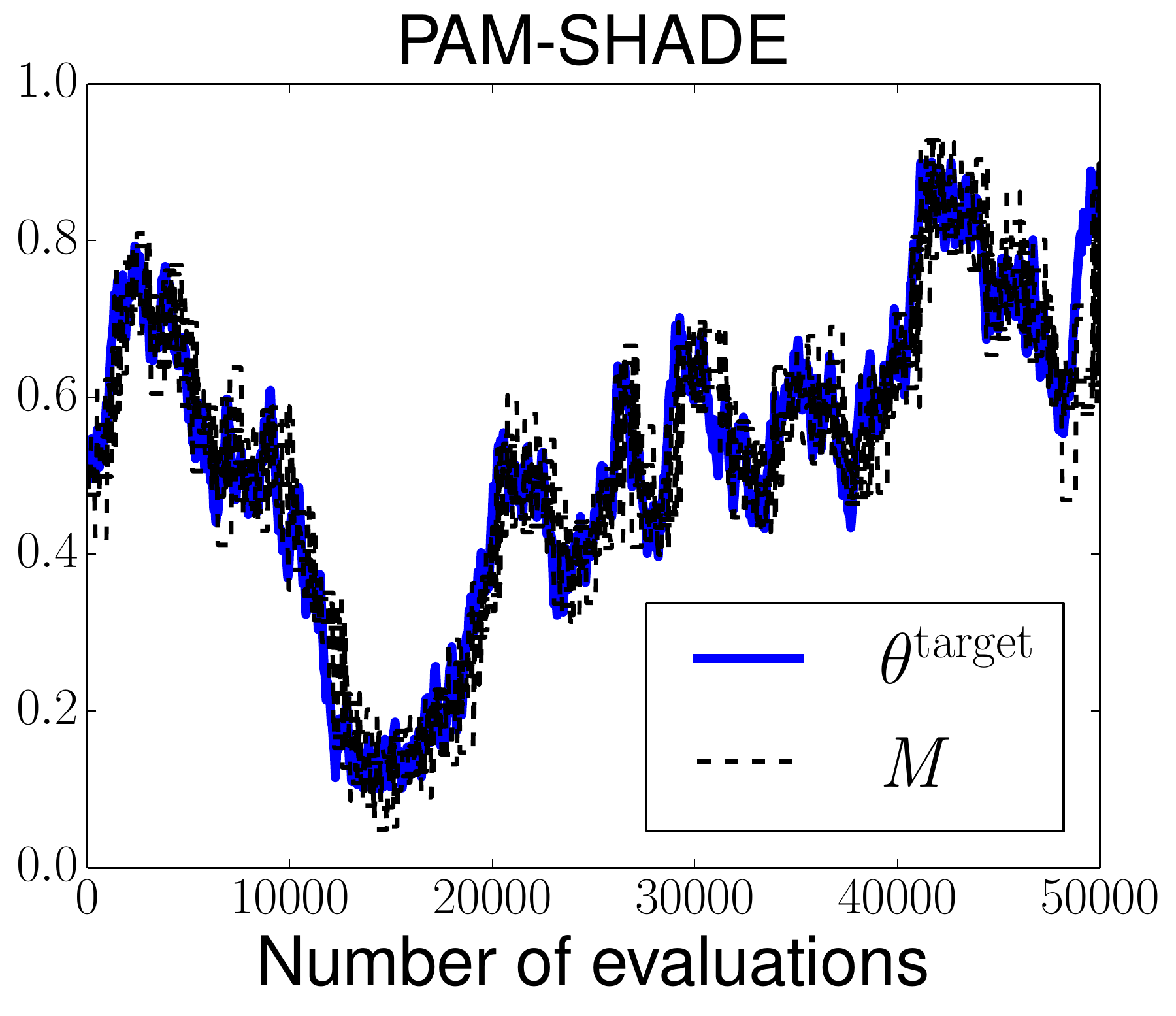}
}
\\[-0.1em]
\subfloat[$g^{\rm ran}$ $(s = 0.1)$]{
\label{subfig:metaparams_s0.1}
\includegraphics[width=\widthvar\textwidth]{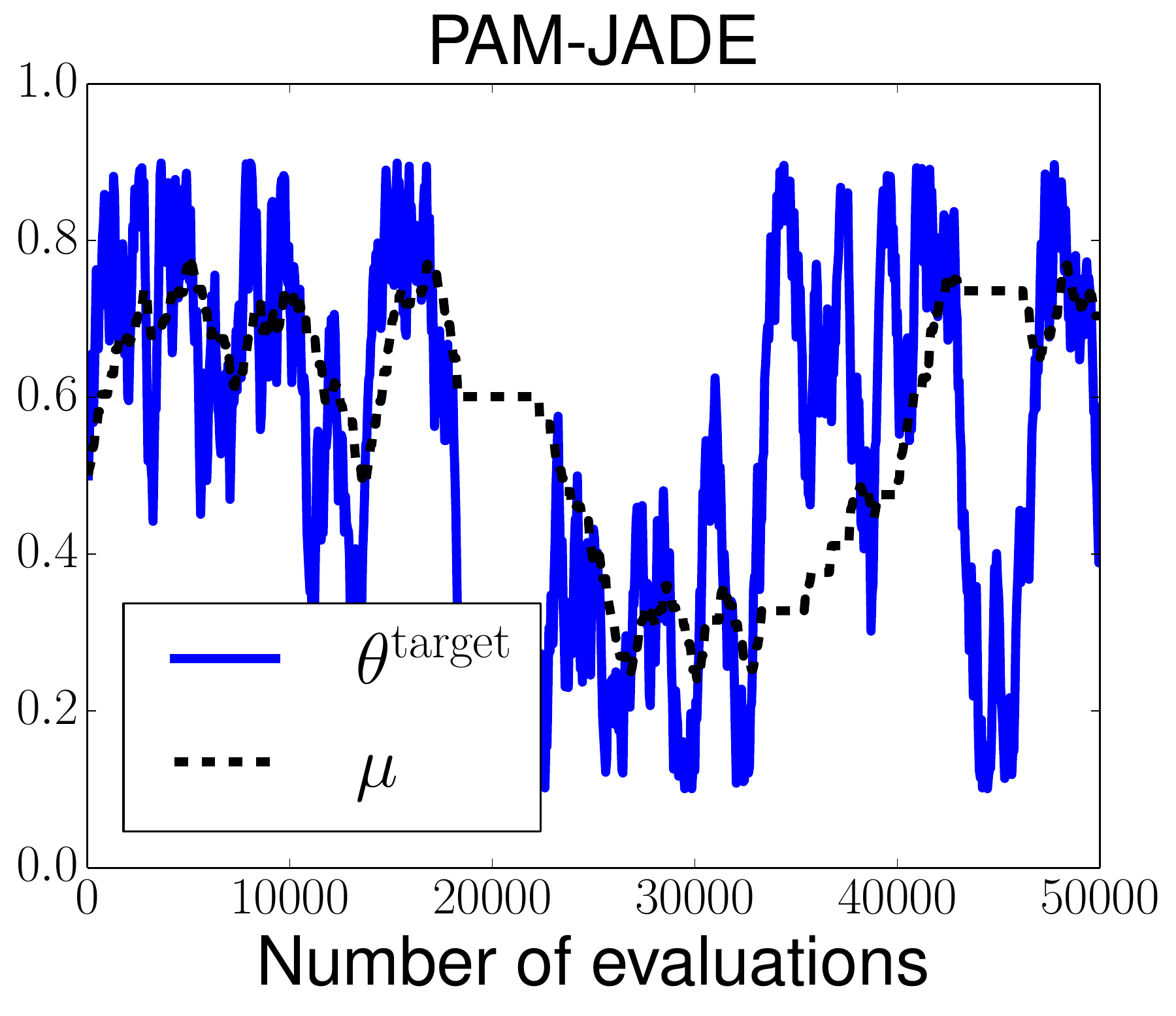}
\includegraphics[width=\widthvar\textwidth]{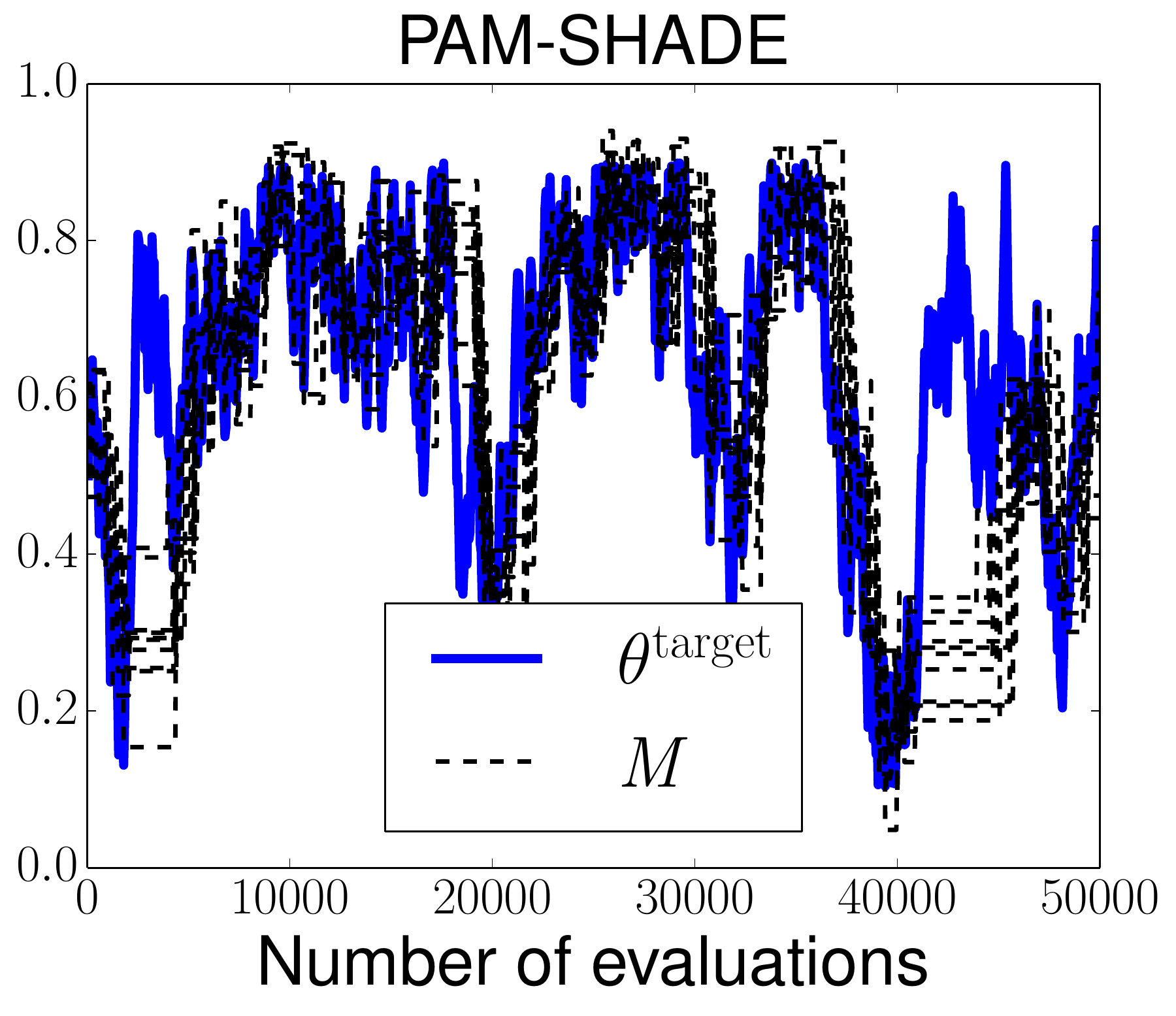}
}
\caption{
\small
Behavior of the meta-parameters ($\mu$ and $\vector{M}$) of PAM-JADE and PAM-SHADE on the TPAM simulation with the various target functions ($p_a^{\rm max} = 0.1$).
For PAM-SHADE, we plot all elements in $\vector{M}$.
Data of a single run with the median $r^{\rm succ}$ value out of the 101 runs are shown.
The comparison on the same $g^{\rm ran}$ instance can be found in Figure S.1 in the supplemental file.
}
\label{fig:metaparams_tpam}
  \end{center}
\end{figure}

The previous subsection presented aggregated target parameter tracking performance over many runs on many settings.
In this section, we take a closer look at tracking behavior of individual runs on given target functions.
Figure \ref{fig:metaparams_tpam} compares how PAM-JADE and PAM-SHADE track each target function during a single run with the median $r^{\rm succ}$ value.
We chose PAM-SHADE and PAM-JADE for comparison because the results in Section  \ref{sec:results_each_tf} showed that these PAMs had good target tracking performance on difficult settings (e.g., for low max acceptance probability $p_a^{\rm max} = 0.1$).

Figures \ref{fig:metaparams_tpam}\subref{subfig:metaparams_linear0.1} and  \subref{subfig:metaparams_omega10} show that when the target parameter values change relatively smoothly, the $\mu$ value for PAM-JADE mostly overlaps the target.
In contrast, PAM-SHADE tracks the target fairly closer while maintaining a broader band of values in its historical memory $\vector{M}$.
In cases where the target values change relatively smoothly and slowly ($g^{\rm lin}$, $g^{\rm sin}$ with $\omega = 10$, $g^{\rm ran}$ with $s \in \{0.01, 0.02, 0.03\}$), it can be seen that PAM-JADE tracks the target more closely than PAM-SHADE.
This illustrates and explains why PAM-JADE exhibited better tracking performance than PAM-SHADE when the target functions were ``easy''.

In contrast, when the target parameter values change rapidly, the $\mu$ values for PAM-JADE clearly fail to track the target, as can bee seen in Figures \ref{fig:metaparams_tpam}\subref{subfig:metaparams_omega40} and  \subref{subfig:metaparams_s0.1}.
However, PAM-SHADE succeeds in tracking the target parameter value fairly well on these difficult tracking problems.
Thus, although the historical memory $\vector{M}$ used by PAM-SHADE prevents perfect tracking of the target parameter values, 
the diversity of values in $\vector{M}$ enables PAM-SHADE to be much more robust than PAM-JADE on rapidly changing target values which are more difficult to track.

Tanabe and Fukunaga conjectured that ``SHADE allows more robust parameter adaptation than JADE'' \cite{TanabeF13}, but this claim was not directly supported either empirically or theoretically, and we know of no work which has directly evaluated the robustness of PAMs.
Our results above provide direct empirical evidence supporting the claim made in \cite{TanabeF13} regarding the comparative robustness of PAM-SHADE.
This shows that TPAM is a powerful technique for analyzing the adaptive behavior of a PAM.

\begin{figure}[t]
  \small
  \newcommand{\widthvar}{0.23}
  \begin{center} 
\includegraphics[width=\widthvar\textwidth]{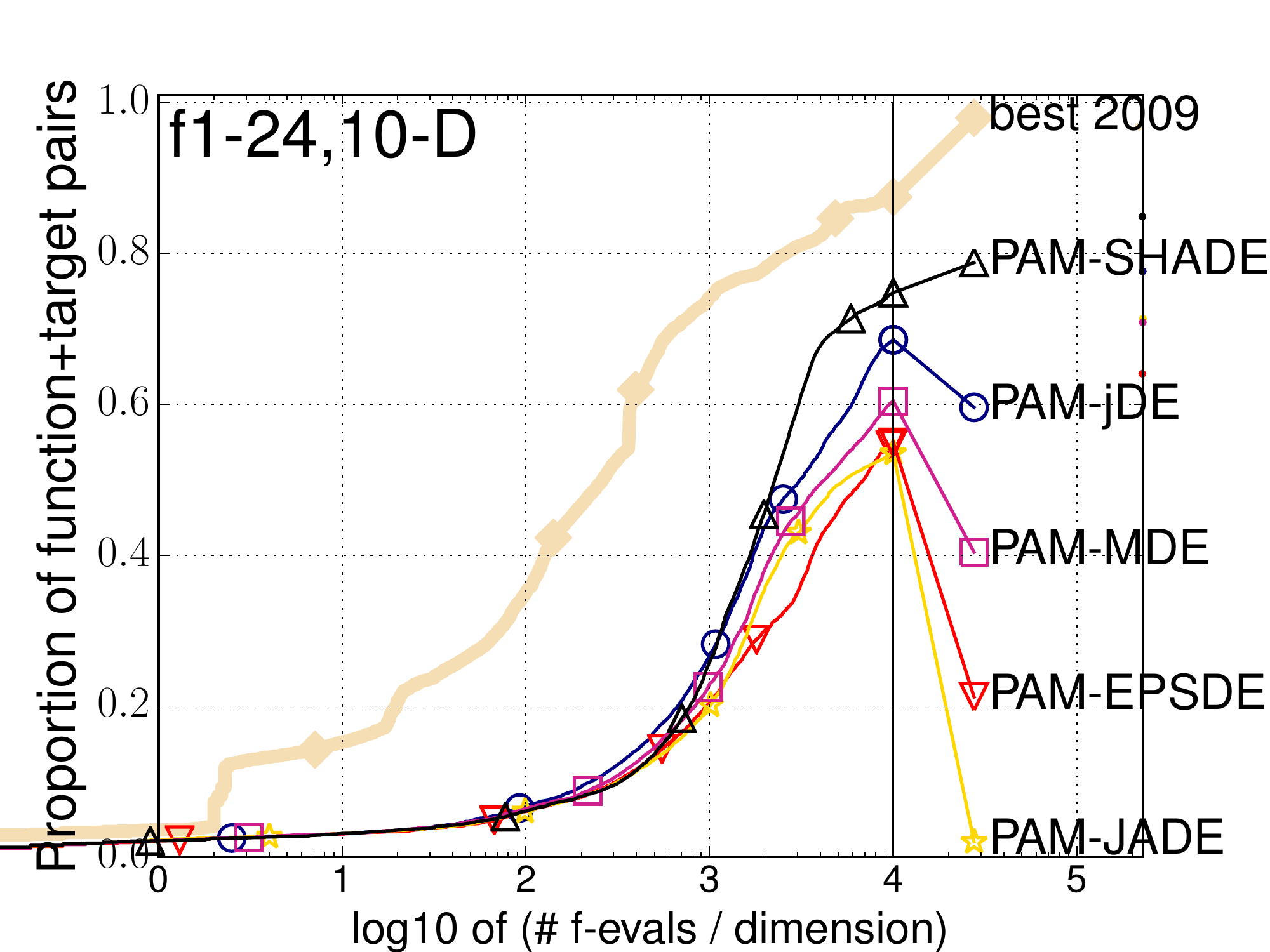}
\includegraphics[width=\widthvar\textwidth]{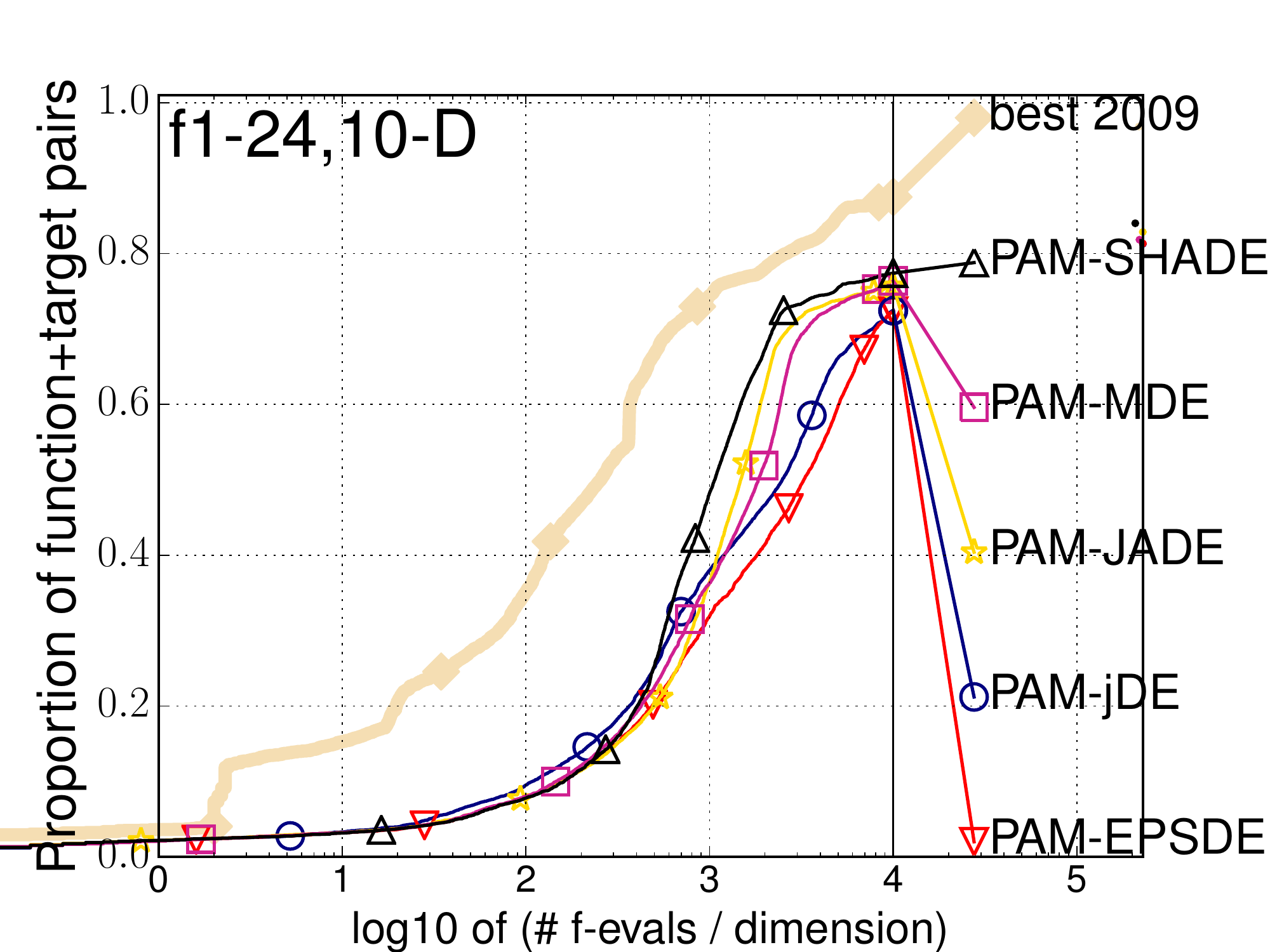}
\caption{
\small
Comparisons of the five PAMs with rand/1/bin and current-to-$p$best/1/bin on the BBOB benchmarks ($D = 10$).
These figures show the bootstrapped Empirical Cumulative Distribution Function (ECDF) of the number of function evaluations (FEvals) divided by dimension for 50 targets in $10^{[-8..2]}$ for 10 dimensional all functions (higher is better).
For details of the ECDF, see a manual of \texttt{COCO software} (\url{http://coco.gforge.inria.fr/}).
}
\label{fig:ade_bbob_result}
  \end{center}
\end{figure}

\subsection{How relevant are the target tracking accuracy of PAMs to the search performance of adaptive DEs?}
\label{sec:bbob_results}

We experimentally verified that the target tracking accuracy measured in these experiments is consistent with the performance of the adaptation mechanisms on standard benchmarks.
We used the noiseless BBOB benchmarks  \cite{hansen2012fun}, comprised of 24 functions $f_1, ..., f_{24}$.
We evaluated all benchmarks with dimensionalities $D \in \{2, 5, 10, 20\}$.
We allocated $10^4 \times D$ function evaluations of each run of each algorithm. The number of trials was 15.
For each PAM, the hyperparameter values were set as recommended in the original papers for each method (see Section \ref{sec:ade} and \ref{sec:experimental-setting}).
Following the work of Po{\v s}{\'{\i}}k and Klema \cite{PosikK12a}, the population size $N$ was set to $5 \times D$ for  $D\geq 5$, and 20 for  $D\leq 3$.
For each method, we evaluated eight different mutation operators (rand/1, rand/2, best/1, best/2, current-to-rand/1, current-to-best/1, current-to-$p$best/1, and rand-to-$p$best/1).
For current-to-$p$best/1 and rand-to-$p$best/1, the control parameters were set to $p = 0.05$ and $|\vector{A}| = N$ \cite{ZhangS09}.
We evaluated both binomial crossover and Shuffled Exponential Crossover (SEC) \cite{PriceSL05,TanabeF14PPSN}.
Since the BBOB benchmark set recommends the use of restart strategies, we used the restart strategy of \cite{ZhabitskyZ13}.

Figure \ref{fig:ade_bbob_result} shows the results for DE using each of the five PAMs on the 10-dimensional BBOB benchmarks ($f_1 \sim f_{24}$) using rand/1 and current-to-$p$best/1 mutation and binomial crossover.
The results for other operators and other dimensions are shown in Figures S.2 $\sim$ S.5 in the supplementary file.
The results on the BBOB benchmarks show that adaptive DE algorithms using PAM-SHADE perform well overall.
This is consistent with the results in Section \ref{sec:results_each_tf}, which showed that PAM-SHADE was able to track target parameter values better than other PAMs when on difficult target functions ($g^{\rm sin}$ and $g^{\rm ran}$ with rapidly varying target parameters).
This suggests that target function tracking performance by the PAM in the TPAM model is 
correlated with search performance of DE using that PAM on standard benchmark functions, and 
target tracking results in the TPAM model can yield insights which are relevant to search algorithm performance.

\section{Conclusion}
\label{sec:conclusion}

This paper explored the question: how can we define and evaluate ``control parameter adaptation'' in adaptive DE.
We proposed a novel framework, TPAM, which evaluates the tracking performance of PAMs with respect to a given target function.
While previous analytical studies on PAMs (e.g., \cite{BrestZBGZ08,ZielinskiWL08,ZhangS09,DrozdikAAT15,SeguraCSL14,ZamudaB15}) have been limited to qualitative discussions,  TPAM enables quantitative comparison of the control parameter adaptation in PAMs. To our knowledge, this is the 
 first quantitative investigation of the parameter adaptation ability of PAMs. % in the research field of EAs. % claiming that this is the first for all EAs is very risky
We evaluated the five PAMs (PAM-jDE, PAM-JADE, PAM-EPSDE, PAM-MDE, PAM-SHADE) of typical adaptive DEs \cite{BrestGBMZ06,ZhangS09,MallipeddiSPT11,IslamDGRS12,TanabeF13}
using TPAM simulations using three target functions ($g^{\rm lin}$, $g^{\rm sin}$, and $g^{\rm ran}$) .
The simulation results showed that the proposed TPAM framework can provide important insights on PAMs.
We also verified that the results of PAMs obtained by the TPAM simulation is mostly consistent with the traditional benchmark methodology using the BBOB benchmarks \cite{hansen2012fun}.
Overall, we conclude that the TPAM is a novel, promising simulation framework for analyzing PAMs in adaptive DE. %as well as EAs. XXother EAs are mentioned in the next paragraph

We believe that the proposed TPAM framework can be applied to analysis of PAMs in other EAs, such as step size adaptation methods in ESs \cite{HansenAA14} and adaptive operator selection methods in GAs with deterministic replacement policies \cite{FialhoSS09}.  This is a direction for future work.
The TPAM framework evaluates only the tracking performance of PAMs, and thus other important aspects of PAM behavior, such as parameter diversity, are not evaluated.
Future work will investigate simulation-based frameworks for evaluating other aspects of PAM behavior,
as well as an
unified, systematic simulation framework (including TPAM) for analyzing the various aspects of PAM behavior.

\bibliographystyle{ACM-Reference-Format}
\bibliography{reference}  % sigproc.bib is the name of the Bibliography in this case

\end{document}